\documentclass[11pt,letterpaper,logo]{mystyle}
\usepackage{iftex}
\ifXeTeX
  \microtypesetup{tracking=false}
\fi
\usepackage{natbib}
\usepackage{float}
\usepackage{fontawesome5}
\usetikzlibrary{arrows.meta,positioning,fit,calc,shapes.geometric,patterns,decorations.pathreplacing,backgrounds}

\runningtitle{RSIGym}
\title{RSIGym: A Flexible Environment for Recursive Self-Improvement}
\author{Fanqing Meng$^{1,2,*}$, Lingxiao Du$^{1,2,*}$, Haocheng Lu$^{1}$, Qiguang Chen$^{1}$\authorcr
Ziqi Zhao$^{1}$, Zijian Wu$^{2}$, Jiayuan Zhuo$^{1}$, Mengkang Hu$^{1,\dagger}$\authorcr
Michael Qizhe Shieh$^{1,2,\dagger}$\\
{\small $^{1}$Evolvent AI \quad $^{2}$National University of Singapore}\\
{\small $^{*}$Equal contribution. $^{\dagger}$Corresponding authors.}\\
{\small \href{https://github.com/evolvent-ai/RSIGym}{\faGithub\ RSIGym}
\quad \href{https://rsi-index.ai/}{\faGlobe\ Website}}}
\date{}

\newcommand{\rsigym}{\textsc{RSIGym}}
\newcommand{\rsindex}{RSI-Index}

\begin{document}

\begin{abstract}
Recursive self-improvement requires carrying accepted changes into later improvement cycles, while studying agent-proposed changes also requires substantial research infrastructure. Existing settings often leave agents to rebuild routine infrastructure or restrict exploration to individual components.
We introduce \rsigym{}, an agent-native research environment based on \emph{Everything as a Service} (EaaS). \rsigym{} exposes training, inference, rollout, evaluation, and sandbox execution through reusable services, with shared budget and permission controls supporting Data, Harness, and Joint improvement tracks.
This design enables agents to investigate individual interventions and jointly optimize data, training settings, and execution harnesses within the same environment.
We define \rsindex{} as the mean fraction of the remaining performance gap closed across five benchmarks covering software engineering, terminal interaction, mathematics, scientific reasoning, and skill-based tasks.
Comparing six frontier research models in independent Joint runs, Opus 5 achieves the highest \rsindex{} of 0.4809 under a \$500 platform-service budget per benchmark run.
Its selected systems improve all five benchmarks, raising SWE-bench Verified from 17.67\% to 50.33\% and AIME from 31.67\% to 97.78\%.
Additional experiments examine DSH-harness refinement, budget variation, and restricted network access, while recorded trajectories reveal how agents diagnose failures and select candidates.
We open-source the full \rsigym{} codebase and results to support reproducibility and further research.

\end{abstract}

\maketitle
\begingroup
\renewcommand{\thefootnote}{\fnsymbol{footnote}}
\footnotetext[2]{Corresponding authors: \href{mailto:mkhu@evolvent.co}{\texttt{mkhu@evolvent.co}} and \href{mailto:qizhex@nus.edu.sg}{\texttt{qizhex@nus.edu.sg}}.}
\endgroup
\thispagestyle{firststyle}

\section{Introduction}
\label{sec:introduction}

Following a recent survey, we use recursive self-improvement (RSI) for a process in which accepted changes are carried into later improvement cycles; recursive progress further requires that inherited changes improve how subsequent improvements are produced \citep{li2026towards}. Automated scientific-research systems such as The AI Scientist generate ideas, run experiments, and analyze results \citep{lu2024aiscientist,chen2025ai4research}. In the improvement loop studied here, an agent identifies a behavioral bottleneck, proposes an intervention, and tests whether the target system improves. Yet existing environments often leave routine training and inference infrastructure for the research agent to assemble. For example, PostTrainBench gives agents an H100 GPU and internet access but no starter code, training data, or hyperparameter configuration, so agents must build the training pipeline themselves \citep{rank2026posttrainbench}. Reusable services can remove this repeated setup work and leave agents more of their time and compute budget for investigating the target system.

Reducing this engineering burden must go hand in hand with preserving the agent's freedom to explore how different components contribute to improvement. Existing benchmarks establish useful, controlled settings by focusing on particular parts of the workflow (Table~\ref{tab:positioning}): optimizing bounded post-training runs and training algorithms \citep{rank2026posttrainbench,chi2026ai4ai}, revising training-data strategies under a shared training stack \citep{meng2026rsibenchdata}, or searching harnesses around fixed actor weights \citep{sleiman2026autoharnessbench}. These settings help study individual interventions, while a broader RSI environment must also support their interactions. Training data and hyperparameters determine the learned model, and the harness determines how that model reasons, uses tools, and acts in a task. Studying their \emph{co-evolution} therefore calls for reusable infrastructure that agents can invoke directly, together with a configurable improvement space. The same environment should support controlled studies of individual components and joint optimization of data, training settings, and harnesses.

We bring these two requirements together in \rsigym{}, an agent-native research environment built on \emph{Everything as a Service} (EaaS). Research agents work in lightweight, CPU-only containers and conduct experiments through five reusable services. A \emph{Train Server} executes LoRA fine-tuning through Tinker with optional custom losses, and a \emph{Model Server} exposes base models and trained checkpoints through an OpenAI-compatible gateway. A \emph{Rollout Server} provides frontier models for data generation and judging, while a \emph{Benchmark Server} evaluates model--harness pairs through Harbor. A separate \emph{Sandbox Service} provides E2B cloud sandboxes for data generation, testing, and other computation. Together, these services support a continuous research loop: the agent proposes a change, trains or revises a candidate, evaluates it, and uses the returned evidence to decide what to try next. Familiar interfaces, service skill documents, and asynchronous execution make this loop directly accessible to research agents.

The scope of this research loop is controlled through permissions and budgets, allowing the same services to support different experimental questions. A shared authorization service checks every call against the run's permissions and remaining budget and records its cost. In the Data track, the agent changes training data while the training recipe and harness remain fixed. In the Harness track, it revises the execution code with model weights held fixed. The Joint track opens data, training settings, and harness code together. These settings make it possible to study individual interventions and their co-evolution within a common infrastructure. Our experiments measure target-system gains within individual budgeted research runs. Whether the resulting systems improve subsequent research cycles remains to be evaluated.

To measure what agents achieve within these improvement spaces, we introduce \rsindex{} across SWE-bench Verified, Terminal-Bench 2.0, AIME, GPQA Diamond, and SkillsBench. The index averages the fraction of the remaining performance gap closed on each benchmark, giving all five domains equal weight. In the Joint track, six frontier research models start from the same base model and minimal harness. With a \$500 platform-service budget per benchmark run, Opus 5 achieves the highest \rsindex{} of 0.4809. Its selected systems improve all five benchmarks, raising SWE-bench Verified from 17.67\% to 50.33\% and AIME from 31.67\% to 97.78\%.

The environment's configurable controls also let us examine how outcomes vary with the starting harness, available budget, and access to external data. With Qwen3.6-35B-A3B-Instruct weights fixed, autonomous revisions to the existing DSH-harness raise Terminal-Bench performance from 30.34\% to 40.82\%, increasing the number of tasks solved on all three attempts from 15 to 26. Raising the service budget from \$500 to \$1,000 improves three of four tested research-model--benchmark pairs. An additional Data-track run disables public network access to examine research under network isolation. The recorded trajectories connect these outcomes to the agents' experiments, showing how they diagnose failures, compare candidates, and select final systems. \rsigym{} thus supports studying both the improvements agents produce and the research decisions that lead to them.

Our contributions are threefold:
\begin{itemize}[leftmargin=*,topsep=0pt]
    \item \textbf{An Agent-Native RSI Environment.} We introduce \rsigym{}, which combines reusable research services with shared budget and permission controls to support both component-level interventions and joint optimization.
    \item \textbf{Cross-Domain Evaluation and Empirical Analysis.} We define \rsindex{}, compare six frontier research models across five benchmarks, and analyze how research decisions and experimental settings shape improvement.
    \item \textbf{Open Code and Experimental Records.} We release the full codebase, experiment details, and evaluation results to support reproduction and further research.
\end{itemize}

\begin{figure}[t]
\centering
\definecolor{SvcTrain}{HTML}{3370FF}
\definecolor{SvcInfer}{HTML}{0F9D9A}
\definecolor{SvcRollout}{HTML}{7B61FF}
\definecolor{SvcEval}{HTML}{E0781A}
\definecolor{SvcSandbox}{HTML}{4E9A3E}
\begin{tikzpicture}[
    scale=0.85, every node/.style={transform shape},
    font=\footnotesize\sffamily,
    >={Stealth[length=5.5pt, width=4.5pt]},
    stagebox/.style={draw=EvolventBorder, fill=white, rounded corners=3pt, line width=0.7pt,
        minimum width=3.3cm, minimum height=1.2cm, text width=3.1cm, align=center, inner sep=3pt},
    chip/.style={rounded corners=2pt, inner xsep=2.5pt, inner ysep=1.5pt,
        minimum width=0pt, minimum height=0pt, text width=, align=center,
        font=\scriptsize\sffamily\bfseries, line width=0.5pt},
    cyc/.style={->, draw=EvolventInk, line width=0.9pt},
    head/.style={font=\footnotesize\bfseries\sffamily, color=EvolventInk, anchor=north west},
    note/.style={font=\scriptsize\sffamily, color=EvolventGray},
    panel/.style={draw=EvolventBorder, fill=EvolventLightGray, rounded corners=5pt, line width=0.6pt},
    provider/.style={chip, draw=EvolventGray, fill=white, text=EvolventInk, font=\scriptsize\sffamily},
    svcbox/.style={rounded corners=3pt, line width=0.8pt, minimum width=3.05cm,
        minimum height=1.75cm, text width=2.85cm, align=center, inner sep=3pt},
    authlink/.style={<->, >={Stealth[length=4pt, width=3.5pt]}, draw=EvolventGray,
        dashed, line width=0.6pt},
]
\newcommand{\svc}[2]{\tikz[baseline=(c.base)]\node[chip, draw=#1, fill=#1!12, text=#1!70!black] (c) {#2};}

\draw[panel] (0, 3.0) rectangle (11.7, -3.2);
\node[head] at (0.15, 2.9) {\faSync\ \ Research loop};

\node[draw=EvolventInk, fill=EvolventTagBg!60, rounded corners=4pt, line width=0.8pt,
    minimum width=2.9cm, text width=2.75cm, align=center, inner sep=4pt] (agent) at (5.85, -0.15)
    {{\Large\faUserCog}\\[2pt]\textbf{Research agent $\pi$}\\[1pt]
     {\scriptsize\color{EvolventGray}CPU-only container}};

\node[stagebox] (revise) at (5.85, 1.85)
    {{\faLightbulb[regular]}\ \textbf{Analyze \& revise}\\[1pt]
     {\scriptsize\color{EvolventGray}data, config, harness}};
\node[stagebox] (gen) at (9.75, -0.15)
    {{\color{SvcRollout}\faDatabase}\ \textbf{Generate data}\\[3pt]
     \svc{SvcRollout}{Rollout}\,\svc{SvcSandbox}{Sandbox}};
\node[stagebox] (train) at (5.85, -2.2)
    {{\color{SvcTrain}\faMicrochip}\ \textbf{Train}\\[3pt]
     \svc{SvcTrain}{Training}};
\node[stagebox] (eval) at (1.95, -0.15)
    {{\color{SvcEval}\faClipboardCheck}\ \textbf{Evaluate}\\[3pt]
     \svc{SvcEval}{Evaluation}\,\svc{SvcInfer}{Inference}};

\draw[cyc] (revise.east) to[out=0, in=90] (gen.north);
\draw[cyc] (gen.south)   to[out=-90, in=0] node[pos=0.35, right, note, xshift=2pt] {data $D$} (train.east);
\draw[cyc] (train.west)  to[out=180, in=-90] node[pos=0.65, left, note, xshift=-2pt] {checkpoint $M$} (eval.south);
\draw[cyc] (eval.north)  to[out=90, in=180] node[pos=0.35, left, note, align=right, xshift=-2pt] {scores,\\trajectories} (revise.west);

\draw[panel] (12.3, 3.0) rectangle (17.1, -3.2);
\node[head] at (12.45, 2.9) {\faFlagCheckered\ \ Final evaluation};
\node[draw=EvolventInk, fill=white, rounded corners=2pt, inner sep=4pt] (sub) at (14.7, 1.35)
    {$(M^{\star},h^{\star})$};
\node[chip, draw=SvcEval, fill=SvcEval!12, text=SvcEval!70!black,
    font=\scriptsize\sffamily, inner sep=4pt] (off) at (14.7, -0.2)
    {\textbf{Evaluation}\\fixed protocol $\mathcal{E}$};
\node[draw=EvolventInk, fill=EvolventTagBg, rounded corners=2pt, inner sep=4pt] (score) at (14.7, -1.75)
    {$s(\pi)$};
\draw[->, line width=0.7pt] (sub) -- (off);
\draw[->, line width=0.7pt] (off) -- (score);

\draw[->, line width=0.9pt, draw=EvolventInk, rounded corners=5pt]
    (revise.north) -- (5.85, 3.35) -- node[above, note] {submit the final system} (12.0, 3.35)
    |- (sub.west);

\draw[panel] (0, -3.75) rectangle (17.1, -8.05);
\node[head] at (0.15, -3.85) {\faCubes\ \ Services};

\foreach \name/\col/\icon/\title/\desc/\prov [count=\i] in {
    strain/SvcTrain/\faMicrochip/Training/{LoRA fine-tuning,\\returns a checkpoint}/Tinker,
    sinfer/SvcInfer/\faBolt/Inference/{OpenAI API over\\trained checkpoints}/Tinker,
    sroll/SvcRollout/\faComments/Rollout/{OpenAI API over\\frontier models}/LiteLLM,
    seval/SvcEval/\faClipboardCheck/Evaluation/{benchmark jobs with\\the agent's harness}/Harbor,
    ssand/SvcSandbox/\faCube/Sandbox/{isolated cloud\\sandboxes}/E2B} {
  \node[svcbox, draw=\col, fill=\col!7] (\name) at ({0.35 + 1.525 + (\i-1)*3.35}, -5.3)
      {{\color{\col}\icon}\ \textbf{\color{\col!75!black}\title}\\[1pt]
       {\scriptsize\color{EvolventInk}\desc}\\[3pt]
       \tikz[baseline=(c.base)]\node[provider] (c) {\prov};};
}

\node[draw=EvolventInk, fill=white, rounded corners=3pt, line width=0.7pt, dashed,
    minimum width=16.4cm, minimum height=0.7cm, align=center] (auth) at (8.55, -7.35)
    {{\faKey}\ \textbf{Authorization}\quad
     {\scriptsize\color{EvolventGray}one key per run $\cdot$ permission check before each call $\cdot$ shared budget billed after each call}};
\foreach \name in {strain, sinfer, sroll, seval, ssand}
  \draw[authlink] (\name.south) -- (\name.south |- auth.north);
\node[note, anchor=west] at ([xshift=3pt]strain.south |- 0,-6.55) {verify / bill};

\draw[->, line width=2.6pt, draw=EvolventAccent!45] (5.85, -3.25) -- (5.85, -3.7);
\node[note, anchor=west] at (6.0, -3.47) {service calls};
\draw[->, line width=2.6pt, draw=EvolventAccent!45] (14.7, -3.25) -- (14.7, -3.7);
\node[note, anchor=west] at (14.85, -3.47) {service calls};
\end{tikzpicture}
\caption{\textbf{Overview of \rsigym{}.} Inside a lightweight container, the research agent iterates a research loop: it generates data, trains a checkpoint, evaluates the checkpoint with its harness, and analyzes the results to revise its data, training configuration, or harness. Colored tags mark the services each stage uses. The services themselves, shown at the bottom, run on external providers. All five share one authorization service, which checks every call against the run's platform key and tracks the run's budget (Sections~\ref{sec:budget-control} and~\ref{sec:permission-control}). When the run ends, the agent submits its final system, which the same evaluation service scores under a fixed protocol.}
\label{fig:architecture}
\end{figure}
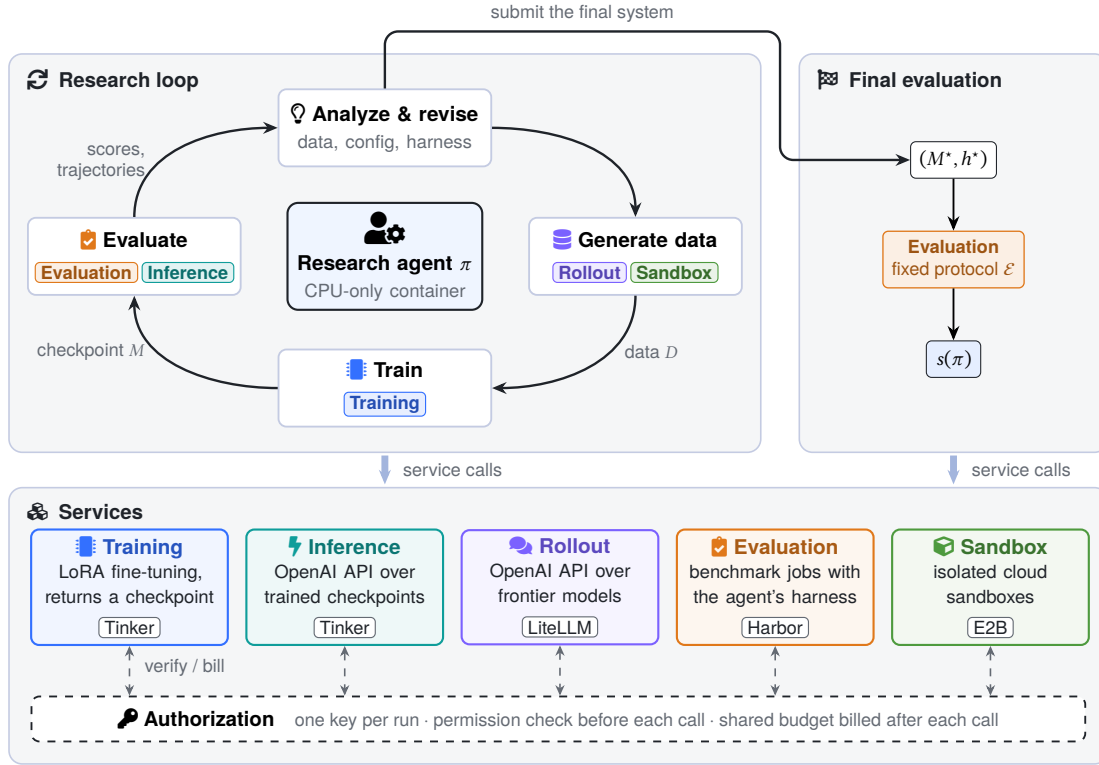

\section{Related Work}
\label{sec:related-work}

\begin{table}[t]
\caption{\textbf{Exploration targets and system designs of benchmarks and environments for AI self-improvement.} A checkmark denotes an explicitly supported exploration direction; a cross denotes a direction outside the main compared setting. System design describes how experimentation and evaluation are organized.}
\label{tab:positioning}
\centering
\footnotesize
\setlength{\tabcolsep}{4pt}
\renewcommand{\arraystretch}{1.25}
\newcommand{\yesmark}{\ding{51}}
\newcommand{\nomark}{\ding{55}}
\begin{tabularx}{\linewidth}{@{}p{0.26\linewidth}>{\centering\arraybackslash}p{0.075\linewidth}>{\centering\arraybackslash}p{0.075\linewidth}>{\centering\arraybackslash}p{0.14\linewidth}X@{}}
\toprule
\textbf{Benchmark / Environment} & \multicolumn{3}{c}{\textbf{RSI Exploration}} & \textbf{System Design} \\
\cmidrule(lr){2-4}
& \shortstack{Model\\Weights} & \shortstack{Harness\\Design} & \shortstack{Algorithm\\Hyperparameters} & \\
\midrule
MLGym\textsuperscript{a} \citep{nathani2025mlgym}
& \yesmark & \nomark & \yesmark & Gym-style container environments \\
PostTrainBench \citep{rank2026posttrainbench}
& \yesmark & \nomark & \yesmark & One H100 GPU; agents build the training pipeline \\
Agent\textsuperscript{2} RL-Bench \citep{chen2026agent2}
& \yesmark & \nomark & \yesmark & Isolated workspaces with a grading API \\
AI4AI-Bench\textsuperscript{a} \citep{chi2026ai4ai}
& \yesmark & \nomark & \yesmark & Code submission, clean reruns, and separate evaluation \\
AutoHarnessBench\textsuperscript{b} \citep{sleiman2026autoharnessbench}
& \nomark & \yesmark & \nomark & Harness search with public feedback and held-out evaluation \\
RSIBench-Data\textsuperscript{c} \citep{meng2026rsibenchdata}
& \yesmark & \nomark & \yesmark & Fixed training, serving, and evaluation services \\
\midrule
\rowcolor{LightBlue}
\textbf{\rsigym{} (ours)}
& \yesmark & \yesmark & \yesmark & \textbf{Everything as a Service} \\
\bottomrule
\end{tabularx}

\vspace{4pt}
\begin{minipage}{\linewidth}
\scriptsize
\textsuperscript{a}Weight optimization applies to learning tasks; these suites also include other task types.
\textsuperscript{b}The main fixed-weight harness tracks are compared; the post-training extension is outside this row's scope.
\textsuperscript{c}Whitelisted training hyperparameters are editable.
Algorithm hyperparameters refer to training settings such as learning rate, batch size, and training steps; decoding and runtime settings belong to harness design.
A cross does not rule out extensions beyond the compared protocol.
\end{minipage}
\end{table}

\paragraph{Recursive Self-Improvement (RSI).}
Automated research systems such as The AI Scientist generate and test scientific ideas \citep{lu2024aiscientist,chen2025ai4research}. A recent survey distinguishes RSI by requiring accepted changes to be retained as the starting state of later improvement cycles; it reserves recursive progress for the stronger case in which inherited changes also improve how subsequent changes are produced \citep{li2026towards}. An earlier formal proposal for self-improvement is Schmidhuber's Gödel machine, which searches for provably beneficial changes to its own program \citep{schmidhuber2007godel}.
Early methods such as AlphaGo Zero self-play and STaR show how systems can improve through self-generated experience \citep{silver2017mastering,zelikman2022star}. The AI Scientist extends automated research to tasks such as coding experiments and analyzing results \citep{lu2024aiscientist}.
Recent benchmarks evaluate specialized slices of this workflow, such as containerized workspaces in MLGym \citep{nathani2025mlgym}, autonomous fine-tuning in PostTrainBench \citep{rank2026posttrainbench}, reinforcement learning in Agent\textsuperscript{2} RL-Bench \citep{chen2026agent2}, and training-algorithm design in AI4AI-Bench \citep{chi2026ai4ai}.\vspace{-8pt}

\paragraph{Post-Training \& Harnessing Optimization.}
DeepSeek-R1 reports that large-scale reinforcement learning can elicit reasoning behavior and improve performance on verifiable reasoning tasks \citep{guo2025deepseekr1}. Agent performance also depends on the tools and runtime that mediate model--environment interaction. AutoHarnessBench studies harness search around fixed model weights, while RSIBench-Data studies data-centered post-training under a shared training and evaluation stack \citep{sleiman2026autoharnessbench,meng2026rsibenchdata}.
These settings isolate complementary research choices: RSIBench-Data holds the post-training stack fixed while agents revise training data; AutoHarnessBench holds actor weights fixed while agents revise the harness.

As formalized in Table~\ref{tab:positioning}, existing environments restrict autonomous exploration to isolated niches: they either optimize model weights while fixing scaffolding, search harnesses around frozen weights, or rely on fragile host-level scripts.
Agent performance depends on both model behavior and the interaction layer. Self-Harness studies trace-driven, model-specific harness changes; HiAgent and SWE-agent examine working-memory management and agent--computer interfaces, respectively \citep{zhang2026selfharness,hu2025hiagent,yang2024sweagent}. These studies motivate evaluating model and harness changes in the same environment, while leaving the effects of their interaction to measurement.
\rsigym{} departs from these paradigms through the \emph{Everything as a Service} (EaaS) architecture, completely decoupling distributed training, inference proxying, and containerized evaluation into hardened microservices under fail-closed budget governance.

\section{System Design}
\label{sec:system-design}

In this section, we formalize the budgeted target-system improvement task supported by \rsigym{} and detail its architecture, organized around the \emph{Everything as a Service} (EaaS) paradigm, unified budget accounting, and automated scientific auditing.

\subsection{Problem Definition}
\label{sec:problem-definition}

\paragraph{Research task.}
\rsigym{} distinguishes the research agent $\pi$, which designs and runs experiments, from the target system $(M,h)$, where $M$ is the model and $h$ is its execution harness. Each experiment specifies an initial system $(M_0,h_0)$, an evaluation protocol $\mathcal{E}$, permitted changes $\mathcal{P}$, and a platform-service budget $\mathcal{B}_{\mathrm{total}}$. The agent's task is to improve the target system's evaluation score within these permissions and the allocated budget.

\paragraph{Permitted changes and research process.}
%
%
Depending on $\mathcal{P}$, the agent may revise training data $D$, training settings $\lambda$, and harness code $h$. Through iterative experiments under the allocated budget, it selects a final model--harness pair:

\begin{equation}
(M^\star,h^\star)
=\pi(M_0,h_0,\mathcal{E};\mathcal{P},\mathcal{B}_{\mathrm{total}}).
\label{eq:research-process}
\end{equation}

Here, $\pi$ denotes the complete research process, including candidate generation, evaluation, and selection. In harness-only experiments, $M^\star=M_0$.

\paragraph{Submission, evaluation, and audit.}
%
%
The final pair $(M^\star,h^\star)$ is represented by a checkpoint identifier and a harness archive, with fixed components supplied by the task. The verifier evaluates this pair under $\mathcal{E}$ and returns the mean task reward. Its score and improvement over the initial system are

\begin{equation}
s(\pi)=\operatorname{Eval}_{\mathcal{E}}(M^\star,h^\star),
\qquad
\Delta(\pi)=s(\pi)-\operatorname{Eval}_{\mathcal{E}}(M_0,h_0).
\label{eq:system-improvement}
\end{equation}

Separate post-run audits assess submission validity through checkpoint checks, training-data overlap analysis, and behavioral review of the research trajectory and submitted artifacts.

\subsection{Everything as a Service}
\label{sec:everything-as-a-service}

Carrying out the process in Section~\ref{sec:problem-definition} requires repeated training runs, model serving, and benchmark evaluation, each of which depends on substantial infrastructure. In \rsigym{}, this infrastructure sits behind services rather than inside the research agent's workspace. The agent works in a lightweight, CPU-only container that holds its harness repository, its data, and its notes. Training, inference, evaluation, and sandbox execution all run remotely, and the agent reaches them through network interfaces (Figure~\ref{fig:architecture}). It does not install GPU drivers, launch distributed training, deploy models, or build evaluation environments. The platform holds the credentials for all underlying providers, while the agent holds a single platform key issued for its run.

Compared with giving each agent a dedicated GPU workspace, this design consumes compute only while a service executes submitted work, rather than leaving it idle while the agent reasons. Work is not bounded by a single machine; an evaluation job, for example, runs many trials concurrently in separate sandboxes. Every research agent also works with the same infrastructure, so differences between agents reflect their research decisions rather than how well they configure an environment.

\paragraph{Services.}
Table~\ref{tab:services} lists the interface of each service, and we describe the role of each below. The \textbf{training service} carries out training proposals generated during the research process in Equation~\eqref{eq:research-process}. The agent submits training data and a training configuration, optionally with a custom loss function, and obtains a checkpoint. The current implementation performs LoRA fine-tuning \citep{hu2022lora} on Tinker \citep{tinker2025}. The \textbf{inference service} makes checkpoints callable. It serves base models and trained checkpoints through an OpenAI-compatible chat completion interface with tool calls, so a checkpoint can be driven by a harness or probed directly by the agent as soon as training finishes, without an explicit deployment step. The \textbf{rollout service} gives access to frontier models through the same interface, routed through LiteLLM \citep{litellm2023}. Their outputs can serve as training data, and the models can also act as judges. The \textbf{evaluation service} scores a model--harness pair, as in Equation~\eqref{eq:system-improvement}. It runs benchmark jobs in the configuration format of Harbor \citep{harbor2026}, with each trial in its own cloud sandbox. A job may carry the agent's own harness as a code archive, which the service installs and runs in every trial against the model endpoint named in the job. The same harness can therefore be paired with the base model or with any trained checkpoint. A job returns aggregate scores, per-task rewards, and the full trajectories and logs. The \textbf{sandbox service} provides general-purpose computation. It offers isolated cloud sandboxes through the E2B SDK \citep{e2b2023}, which the agent can use for data generation, test execution, or other scratch work, and for which it can build custom images.

\begin{table}[t]
\caption{\textbf{The five services of \rsigym{}.} Each service accepts a request from the research agent, performs the work on an underlying provider, and returns the results and artifacts listed here.}
\label{tab:services}
\centering
\footnotesize
\renewcommand{\arraystretch}{1.2}
\begin{tabular}{@{}l@{\hspace{8pt}}l@{\hspace{8pt}}l@{\hspace{8pt}}l@{}}
\toprule
\textbf{Service} & \textbf{Agent submits} & \textbf{Service returns} & \textbf{Backend} \\
\midrule
\textbf{Training}   & Training configuration, dataset, optional loss function & Per-step metrics, checkpoint identifier & Tinker \\
\textbf{Inference}  & Chat completion request to a base model or checkpoint & Completion with tool calls & Tinker \\
\textbf{Rollout}    & Chat completion request to a frontier model & Completion with tool calls & LiteLLM \\
\textbf{Evaluation} & Job configuration, optional harness archive & Scores, per-task rewards, trajectories, logs & Harbor \\
\textbf{Sandbox}    & Sandbox and image requests via the E2B SDK & Isolated sandboxes, custom images & E2B \\
\bottomrule
\end{tabular}
\end{table}

\paragraph{Authorization.}
Behind the five services sits a shared authorization service. It issues a platform key for each run and stores the permissions and budget attached to that key. Every service verifies each call with the authorization service before executing it and reports the cost of the work once it completes. Permissions and spending are therefore managed in one place, so the same services can serve every experiment, with the rules of each experiment carried by its key. Sections~\ref{sec:budget-control} and~\ref{sec:permission-control} describe how the budget and permissions are enforced through this service.

\paragraph{Interface design.}
Five properties shape these interfaces. \emph{(i)~Existing protocols.} Inference and rollout calls use the OpenAI chat completion format, evaluation jobs use Harbor's configuration schema, and sandboxes use the unmodified E2B SDK. These protocols are widely adopted and appear extensively in public code and documentation, so frontier models acting as research agents are already fluent in them and need not learn a platform-specific interface. \emph{(ii)~Asynchronous execution.} Training runs and evaluation jobs, which can take hours, run in the background. Submitting one returns a handle right away, so the agent is never blocked waiting for a result and can run several experiments in parallel, cancelling unpromising ones early. \emph{(iii)~Inspectable evidence.} Each call returns evidence rather than a single number. Training reports its loss curve, and evaluation reports which tasks passed and how the harness behaved on each of them, so the agent can diagnose why a candidate succeeded or failed rather than only whether its score changed. \emph{(iv)~Self-documentation.} Each service comes with a skill document that describes its endpoints, configuration fields, and errors, so any research agent that can read documentation can use the platform without platform-specific training or prompting. \emph{(v)~Recorded execution.} The services keep their own record of the work they execute, independent of what the agent reports. Each training run retains its submitted data and loss function, each evaluation job retains its trajectories and logs, and the authorization service keeps an itemized ledger of costs. These records allow a run to be audited after the fact.

\paragraph{Closing the loop.}
Together, these services support a complete research loop. In each iteration, the agent trains a checkpoint, evaluates it with its harness, reads the resulting trajectories, and revises its data, training configuration, or harness accordingly. When the run ends, the agent submits its selected pair $(M^{\star},h^{\star})$ as a checkpoint identifier and a harness archive. The platform evaluates this pair once through the same evaluation service, with a job configuration fixed in advance by the evaluation protocol $\mathcal{E}$, and the resulting score is $s(\pi)$ in Equation~\eqref{eq:system-improvement}.

\subsection{Budget Control}
\label{sec:budget-control}

Each experiment grants the research agent a budget $\mathcal{B}_{\mathrm{total}}$, and its use of the services is charged against this budget according to the resources consumed. Before executing a call, a service checks the remaining budget with the authorization service, and it records the cost once the work completes. Once the budget is exhausted, the agent can no longer perform any operation that incurs cost, and a training run already in progress is stopped as well. The budget is also visible to the agent. Each call reports its cost, a training run reports its accumulated cost as it progresses, and the agent can query its remaining balance at any time. How to spend the budget therefore becomes part of the research problem, as the agent decides how much to invest in generating data, training candidates, and evaluating them. Because the budget is a parameter of the experiment, varying $\mathcal{B}_{\mathrm{total}}$ while holding $\mathcal{P}$ and $\mathcal{E}$ fixed shows how the results of a research agent scale with the resources available to it.

\subsection{Permission Control}
\label{sec:permission-control}

The permission configuration $\mathcal{P}$ determines what the research agent may change and which resources it may use. Permissions are bound to the platform key of each run and checked on every service call, together with the budget. They are set at two levels. At the service level, each service can be granted or revoked; without the training service, for example, the agent can only improve the harness. Within a granted service, individual settings can be left open, restricted to an allowed set of values, or fixed. This determines, for example, which algorithm hyperparameters $\lambda$ the agent may set, which frontier models it may call through the rollout service, and whether its sandboxes can reach the internet.

Because all experiments share the same services, the scope of an experiment is largely defined by its permissions. A harness-only experiment revokes the training and rollout services, so the model stays fixed and the agent improves only the harness. A data-only experiment grants the training service but fixes all algorithm hyperparameters, so the agent improves the system only through the data it designs. A joint experiment opens the data, the training settings, and the harness together, fixing only the base model. The offline setting is defined in the same way, by revoking internet access. Studying one component alone or several together therefore requires no change to the environment, only to the permissions.

\needspace{10\baselineskip}
\section{Experiments}
\label{sec:experimental-design}
\label{sec:experiments}
\subsection{Experimental Setup}
\label{sec:experimental-setup}
\paragraph{Research agents.}
We evaluate six research agents: Claude Opus 5, Claude Opus 5.5, GPT-6 Astra, GPT-6 Sol, GPT-6.1 Sol, and DeepSeek V4.1 Flash (Opus 5, Opus 5.5, Astra, Sol 6, Sol 6.1, and DeepSeek below). Opus and DeepSeek run in Claude Code; Astra and both Sol versions run in Codex.

\paragraph{Tracks.}
In the \textbf{Joint} track, the main study, the agent may change training data, training settings, and harness. It starts from Qwen3.5-35B-A3B-Base and a minimal model--tool loop\footnote{The minimal harness uses the DeepAgents SDK from LangChain \citep{langchain2026deepagents}, with a short system prompt and default filesystem and shell tools, as in the initial setup of Self-Harness \citep{zhang2026selfharness}. Our implementation uses \texttt{deepagents==0.7.8}.}, with \$500 per benchmark run (Table~\ref{tab:runs}). Two single-freedom tracks, run by Opus 5, isolate one freedom each. \textbf{Data} opens only the training data from the same initial system. \textbf{Harness} opens only the harness and keeps Qwen3.6-35B-A3B-Instruct weights fixed. Four Joint runs repeat SWE and Terminal-Bench with \$1,000 to examine how a larger budget affects performance. Section~\ref{sec:mechanisms} adds three case studies: a Data run without network access, Harness research that starts from an existing framework, DeepSeek Harness (DSH, \texttt{dsh-v0.1.1-rc.1}), and a run in which Qwen3.8-27B improves itself.

\begin{table}[t]
\centering\small
\caption{Runs in this study. Each run is an independent search that submits one system. Budget is the per-run allocation for platform services. The first four rows are reported in this section; the last three are the case studies of Section~\ref{sec:mechanisms}.}
\label{tab:runs}
\begin{tabular}{@{}llllllr@{}}
\toprule
Track & Research agent & Target model & Initial harness & Benchmarks & Budget & Runs \\
\midrule
Joint & all six & Qwen3.5-35B-A3B-Base & minimal & all five & \$500 & 30 \\
Joint & Opus 5, Astra & Qwen3.5-35B-A3B-Base & minimal & SWE, TB2 & \$1,000 & 4 \\
Data & Opus 5 & Qwen3.5-35B-A3B-Base & minimal & SWE, TB2 & \$500 & 2 \\
Harness & Opus 5 & Qwen3.6-35B-A3B-Instruct & minimal & SWE, TB2 & \$500 & 2 \\
\midrule
Data, no network & Opus 5 & Qwen3.5-35B-A3B-Base & minimal  & TB2 & \$500 & 1 \\
Harness & Opus 5 & Qwen3.6-35B-A3B-Instruct & DSH v0.1.1-rc.1 & TB2 & \$500 & 1 \\
Self-improvement & Qwen3.8-27B & Qwen3.8-27B & pi v0.84.2 & Skills & \$500 & 1 \\
\bottomrule
\end{tabular}
\end{table}

\paragraph{Protocol.}
\label{sec:rules-and-audits}
During a run the agent can launch development evaluations and read public benchmark material. The reported score is the final official evaluation of the submitted system. The agent has seen these tasks during development, so scores are not held-out. Every configuration is a single run. Time is wall-clock from run start through the final evaluation, and spend is what the run charges to its platform account for target-model inference, rollouts, and training; research-agent inference and the final evaluation are billed separately. Training examples must be produced by the agent during the run, and the checkpoint must not be trained on evaluation tasks or data derived from them. Our post-run audits check 13-gram overlap with evaluation tasks, checkpoint provenance, and the trajectory and submission through an LLM judge. PostTrainBench also discusses contamination and model-identity checks, but the specific overlap screen and audit procedure reported here are part of this study \citep{rank2026posttrainbench}. All submissions passed the three audits.

\subsection{Benchmarks and Evaluation}
\label{sec:benchmarks-and-evaluation}
\begin{table}[t]
\centering\small
\caption{Evaluation sets. Task counts are the configured subsets; every final job uses three attempts per task.}
\label{tab:benchmarks}
\begin{tabularx}{\linewidth}{@{}lrrX@{}}
\toprule Benchmark & Tasks & Trials & Scoring and configuration \\
\midrule
SWE-bench Verified & 100 & 300 & Fixed seed-23 sample; held-out repository tests \\
Terminal-Bench 2.0 & 89 & 267 & All configured tasks; task verifiers \\
AIME 2024/2025 & 60 & 180 & Correct final integer answer \\
GPQA Diamond & 100 & 300 & Fixed seed-23 sample; correct option \\
SkillsBench & 37 & 111 & Science/office/finance subset; native task reward \\
\bottomrule
\end{tabularx}
\end{table}

We use five benchmarks (Table~\ref{tab:benchmarks}): SWE-bench Verified\footnote{\url{https://www.swebench.com/verified.html}.} \citep{jimenez2024swebench}, Terminal-Bench 2.0 (TB2),\footnote{\url{https://www.tbench.ai/news/announcement-2-0}.} AIME 2024 and 2025 \citep{jia2024aime,mathai2025aime}, GPQA Diamond \citep{rein2024gpqa}, and SkillsBench \citep{li2026skillsbench}.\footnote{For SkillsBench, we use only the science, office, and finance subsets, comprising 37 tasks in total.} The score on benchmark $b$ with $N_b$ tasks is the mean reward over three attempts per task (avg@3):
\begin{equation}
 s_b=\frac{1}{3N_b}\sum_{i=1}^{N_b}\sum_{j=1}^{3}r_{b,i,j}.
 \label{eq:score}
\end{equation}
A verifier reward counts even when the agent records an exception, and a missing reward counts as zero. With $s_b^0$ the score of the initial system and a maximum reward of $1$, the \rsindex{} over $n$ benchmarks is
\begin{equation}
 \text{RSI-Index}=\frac{1}{n}\sum_{b=1}^{n}\frac{s_b\mathbin{\text{\textminus}}s_b^0}{1\mathbin{\text{\textminus}}s_b^0},
\label{eq:rsindex}
\end{equation}
the mean fraction of the available improvement that is closed. It is $0$ for the initial system, $1$ for a perfect system, and negative for a regression, and it weights benchmarks equally regardless of task count. Here $n=5$, and the index is computed from full-precision scores.

\subsection{Results}
\label{sec:results-and-analysis}
\begin{figure}[t]
\centering
\definecolor{rsiInk}{HTML}{1F2328}
\definecolor{rsiMuted}{HTML}{6E7681}
\definecolor{rsiRule}{HTML}{D8DCE2}
\definecolor{rsiGrey}{HTML}{B4BBC5}
\definecolor{rsiTrack}{HTML}{EEF0F3}
\definecolor{rsiBlue}{HTML}{1F6FEB}
\definecolor{rsiRed}{HTML}{C9605D}
\begin{tikzpicture}[x=1cm, y=1cm, line cap=butt,
  every node/.style={inner sep=0pt, outer sep=0pt, text=rsiInk, font=\footnotesize},
  sub/.style={font=\scriptsize, text=rsiMuted},
  head/.style={font=\footnotesize\bfseries},
  axis/.style={draw=rsiInk, line width=0.5pt},
  hair/.style={draw=rsiRule, line width=0.3pt}]
\node[anchor=west, head] at (1.700,0.980) {SWE};
\node[anchor=west, sub] at (1.700,0.600) {initial 0.1767};
\node[anchor=west, head] at (4.060,0.980) {TB2};
\node[anchor=west, sub] at (4.060,0.600) {initial 0.1049};
\node[anchor=west, head] at (6.420,0.980) {AIME};
\node[anchor=west, sub] at (6.420,0.600) {initial 0.3167};
\node[anchor=west, head] at (8.780,0.980) {GPQA};
\node[anchor=west, sub] at (8.780,0.600) {initial 0.5133};
\node[anchor=west, head] at (11.140,0.980) {Skills};
\node[anchor=west, sub] at (11.140,0.600) {initial 0.0165};
\node[anchor=west, head] at (13.600,0.980) {\rsindex{}};
\node[anchor=west, sub] at (13.600,0.600) {mean share};
\node[anchor=east, head] at (16.700,0.980) {Spend};
\node[anchor=east, sub] at (16.700,0.600) {\$};
\node[anchor=east, head] at (17.580,0.980) {Time};
\node[anchor=east, sub] at (17.580,0.600) {hours};
\draw[axis] (0,0.36) -- (17.6,0.36);
\node[anchor=west] at (0.000,-0.120) {Opus 5};
\node[anchor=west] at (1.700,0.040) {0.5033};
\node[anchor=east, sub] at (3.750,0.040) {+40\%};
\fill[rsiTrack] (1.700,-0.420) rectangle (3.750,-0.200);
\fill[rsiGrey] (1.700,-0.420) rectangle (2.513,-0.200);
\node[anchor=west, text=rsiBlue] at (4.060,0.040) {\textbf{0.2697}};
\node[anchor=east, sub] at (6.110,0.040) {+18\%};
\fill[rsiTrack] (4.060,-0.420) rectangle (6.110,-0.200);
\fill[rsiBlue] (4.060,-0.420) rectangle (4.437,-0.200);
\node[anchor=west, text=rsiBlue] at (6.420,0.040) {\textbf{0.9778}};
\node[anchor=east, sub] at (8.470,0.040) {+97\%};
\fill[rsiTrack] (6.420,-0.420) rectangle (8.470,-0.200);
\fill[rsiBlue] (6.420,-0.420) rectangle (8.403,-0.200);
\node[anchor=west] at (8.780,0.040) {0.8333};
\node[anchor=east, sub] at (10.830,0.040) {+66\%};
\fill[rsiTrack] (8.780,-0.420) rectangle (10.830,-0.200);
\fill[rsiGrey] (8.780,-0.420) rectangle (10.128,-0.200);
\node[anchor=west] at (11.140,0.040) {0.2121};
\node[anchor=east, sub] at (13.190,0.040) {+20\%};
\fill[rsiTrack] (11.140,-0.420) rectangle (13.190,-0.200);
\fill[rsiGrey] (11.140,-0.420) rectangle (11.548,-0.200);
\fill[rsiBlue] (13.600,-0.230) rectangle (14.514,-0.010);
\node[anchor=west, text=rsiBlue] at (14.614,-0.120) {\textbf{0.4809}};
\node[anchor=east] at (16.700,-0.120) {1,910.39};
\node[anchor=east] at (17.580,-0.120) {44.92};
\draw[hair] (0,-0.510) -- (17.6,-0.510);
\node[anchor=west] at (0.000,-0.960) {Opus 5.5};
\node[anchor=west] at (1.700,-0.800) {0.5267};
\node[anchor=east, sub] at (3.750,-0.800) {+43\%};
\fill[rsiTrack] (1.700,-1.260) rectangle (3.750,-1.040);
\fill[rsiGrey] (1.700,-1.260) rectangle (2.571,-1.040);
\node[anchor=west] at (4.060,-0.800) {0.2172};
\node[anchor=east, sub] at (6.110,-0.800) {+13\%};
\fill[rsiTrack] (4.060,-1.260) rectangle (6.110,-1.040);
\fill[rsiGrey] (4.060,-1.260) rectangle (4.317,-1.040);
\node[anchor=west] at (6.420,-0.800) {0.9556};
\node[anchor=east, sub] at (8.470,-0.800) {+94\%};
\fill[rsiTrack] (6.420,-1.260) rectangle (8.470,-1.040);
\fill[rsiGrey] (6.420,-1.260) rectangle (8.337,-1.040);
\node[anchor=west, text=rsiBlue] at (8.780,-0.800) {\textbf{0.8533}};
\node[anchor=east, sub] at (10.830,-0.800) {+70\%};
\fill[rsiTrack] (8.780,-1.260) rectangle (10.830,-1.040);
\fill[rsiBlue] (8.780,-1.260) rectangle (10.212,-1.040);
\node[anchor=west] at (11.140,-0.800) {0.0948};
\node[anchor=east, sub] at (13.190,-0.800) {+8\%};
\fill[rsiTrack] (11.140,-1.260) rectangle (13.190,-1.040);
\fill[rsiGrey] (11.140,-1.260) rectangle (11.303,-1.040);
\fill[rsiGrey] (13.600,-1.070) rectangle (14.460,-0.850);
\node[anchor=west] at (14.560,-0.960) {0.4528};
\node[anchor=east] at (16.700,-0.960) {2,083.50};
\node[anchor=east] at (17.580,-0.960) {31.99};
\draw[hair] (0,-1.350) -- (17.6,-1.350);
\node[anchor=west] at (0.000,-1.800) {DeepSeek};
\node[anchor=west, text=rsiBlue] at (1.700,-1.640) {\textbf{0.5333}};
\node[anchor=east, sub] at (3.750,-1.640) {+43\%};
\fill[rsiTrack] (1.700,-2.100) rectangle (3.750,-1.880);
\fill[rsiBlue] (1.700,-2.100) rectangle (2.588,-1.880);
\node[anchor=west] at (4.060,-1.640) {0.2097};
\node[anchor=east, sub] at (6.110,-1.640) {+12\%};
\fill[rsiTrack] (4.060,-2.100) rectangle (6.110,-1.880);
\fill[rsiGrey] (4.060,-2.100) rectangle (4.300,-1.880);
\node[anchor=west] at (6.420,-1.640) {0.8778};
\node[anchor=east, sub] at (8.470,-1.640) {+82\%};
\fill[rsiTrack] (6.420,-2.100) rectangle (8.470,-1.880);
\fill[rsiGrey] (6.420,-2.100) rectangle (8.103,-1.880);
\node[anchor=west] at (8.780,-1.640) {0.8333};
\node[anchor=east, sub] at (10.830,-1.640) {+66\%};
\fill[rsiTrack] (8.780,-2.100) rectangle (10.830,-1.880);
\fill[rsiGrey] (8.780,-2.100) rectangle (10.128,-1.880);
\node[anchor=west] at (11.140,-1.640) {0.1568};
\node[anchor=east, sub] at (13.190,-1.640) {+14\%};
\fill[rsiTrack] (11.140,-2.100) rectangle (13.190,-1.880);
\fill[rsiGrey] (11.140,-2.100) rectangle (11.432,-1.880);
\fill[rsiGrey] (13.600,-1.910) rectangle (14.425,-1.690);
\node[anchor=west] at (14.525,-1.800) {0.4343};
\node[anchor=east] at (16.700,-1.800) {1,095.60};
\node[anchor=east] at (17.580,-1.800) {39.19};
\draw[hair] (0,-2.190) -- (17.6,-2.190);
\node[anchor=west] at (0.000,-2.640) {Astra};
\node[anchor=west] at (1.700,-2.480) {0.3567};
\node[anchor=east, sub] at (3.750,-2.480) {+22\%};
\fill[rsiTrack] (1.700,-2.940) rectangle (3.750,-2.720);
\fill[rsiGrey] (1.700,-2.940) rectangle (2.148,-2.720);
\node[anchor=west] at (4.060,-2.480) {0.2210};
\node[anchor=east, sub] at (6.110,-2.480) {+13\%};
\fill[rsiTrack] (4.060,-2.940) rectangle (6.110,-2.720);
\fill[rsiGrey] (4.060,-2.940) rectangle (4.326,-2.720);
\node[anchor=west] at (6.420,-2.480) {0.8833};
\node[anchor=east, sub] at (8.470,-2.480) {+83\%};
\fill[rsiTrack] (6.420,-2.940) rectangle (8.470,-2.720);
\fill[rsiGrey] (6.420,-2.940) rectangle (8.120,-2.720);
\node[anchor=west] at (8.780,-2.480) {0.8433};
\node[anchor=east, sub] at (10.830,-2.480) {+68\%};
\fill[rsiTrack] (8.780,-2.940) rectangle (10.830,-2.720);
\fill[rsiGrey] (8.780,-2.940) rectangle (10.170,-2.720);
\node[anchor=west] at (11.140,-2.480) {0.2044};
\node[anchor=east, sub] at (13.190,-2.480) {+19\%};
\fill[rsiTrack] (11.140,-2.940) rectangle (13.190,-2.720);
\fill[rsiGrey] (11.140,-2.940) rectangle (11.532,-2.720);
\fill[rsiGrey] (13.600,-2.750) rectangle (14.378,-2.530);
\node[anchor=west] at (14.478,-2.640) {0.4093};
\node[anchor=east] at (16.700,-2.640) {1,342.53};
\node[anchor=east] at (17.580,-2.640) {19.33};
\draw[hair] (0,-3.030) -- (17.6,-3.030);
\node[anchor=west] at (0.000,-3.480) {Sol 6.1};
\node[anchor=west] at (1.700,-3.320) {0.1867};
\node[anchor=east, sub] at (3.750,-3.320) {+1\%};
\fill[rsiTrack] (1.700,-3.780) rectangle (3.750,-3.560);
\fill[rsiGrey] (1.700,-3.780) rectangle (1.725,-3.560);
\node[anchor=west] at (4.060,-3.320) {0.2210};
\node[anchor=east, sub] at (6.110,-3.320) {+13\%};
\fill[rsiTrack] (4.060,-3.780) rectangle (6.110,-3.560);
\fill[rsiGrey] (4.060,-3.780) rectangle (4.326,-3.560);
\node[anchor=west] at (6.420,-3.320) {0.8889};
\node[anchor=east, sub] at (8.470,-3.320) {+84\%};
\fill[rsiTrack] (6.420,-3.780) rectangle (8.470,-3.560);
\fill[rsiGrey] (6.420,-3.780) rectangle (8.137,-3.560);
\node[anchor=west] at (8.780,-3.320) {0.8100};
\node[anchor=east, sub] at (10.830,-3.320) {+61\%};
\fill[rsiTrack] (8.780,-3.780) rectangle (10.830,-3.560);
\fill[rsiGrey] (8.780,-3.780) rectangle (10.030,-3.560);
\node[anchor=west] at (11.140,-3.320) {0.0870};
\node[anchor=east, sub] at (13.190,-3.320) {+7\%};
\fill[rsiTrack] (11.140,-3.780) rectangle (13.190,-3.560);
\fill[rsiGrey] (11.140,-3.780) rectangle (11.287,-3.560);
\fill[rsiGrey] (13.600,-3.590) rectangle (14.231,-3.370);
\node[anchor=west] at (14.331,-3.480) {0.3321};
\node[anchor=east] at (16.700,-3.480) {861.75};
\node[anchor=east] at (17.580,-3.480) {17.55};
\draw[hair] (0,-3.870) -- (17.6,-3.870);
\node[anchor=west] at (0.000,-4.320) {Sol 6};
\node[anchor=west] at (1.700,-4.160) {0.1967};
\node[anchor=east, sub] at (3.750,-4.160) {+2\%};
\fill[rsiTrack] (1.700,-4.620) rectangle (3.750,-4.400);
\fill[rsiGrey] (1.700,-4.620) rectangle (1.750,-4.400);
\node[anchor=west] at (4.060,-4.160) {0.1760};
\node[anchor=east, sub] at (6.110,-4.160) {+8\%};
\fill[rsiTrack] (4.060,-4.620) rectangle (6.110,-4.400);
\fill[rsiGrey] (4.060,-4.620) rectangle (4.223,-4.400);
\node[anchor=west] at (6.420,-4.160) {0.8278};
\node[anchor=east, sub] at (8.470,-4.160) {+75\%};
\fill[rsiTrack] (6.420,-4.620) rectangle (8.470,-4.400);
\fill[rsiGrey] (6.420,-4.620) rectangle (7.953,-4.400);
\node[anchor=west, text=rsiRed] at (8.780,-4.160) {0.5033};
\node[anchor=east, sub] at (10.830,-4.160) {\textminus{}2\%};
\fill[rsiTrack] (8.780,-4.620) rectangle (10.830,-4.400);
\fill[rsiRed] (8.780,-4.620) rectangle (8.840,-4.400);
\node[anchor=west, text=rsiBlue] at (11.140,-4.160) {\textbf{0.2214}};
\node[anchor=east, sub] at (13.190,-4.160) {+21\%};
\fill[rsiTrack] (11.140,-4.620) rectangle (13.190,-4.400);
\fill[rsiBlue] (11.140,-4.620) rectangle (11.567,-4.400);
\fill[rsiGrey] (13.600,-4.430) rectangle (13.995,-4.210);
\node[anchor=west] at (14.095,-4.320) {0.2079};
\node[anchor=east] at (16.700,-4.320) {161.73};
\node[anchor=east] at (17.580,-4.320) {16.03};
\draw[axis] (0,-4.820) -- (17.6,-4.820);
\end{tikzpicture}
\caption{\textbf{Joint track at \$500 per run: all six research agents improve Qwen3.5-35B-A3B-Base, and no agent is best everywhere.} Each cell gives the official score and, as a bar and a percentage, the share of the gap to a perfect score that is closed. Blue marks the best value in a column and red a score below the initial system. \rsindex{} is the mean of the five shares. Spend and time are summed over an agent's five independent runs; time includes final official evaluation.}
\label{fig:joint}
\label{tab:joint}\label{tab:main-results}\label{tab:joint-resources}
\end{figure}

\paragraph{Joint: all six agents improve the target.}
Figure~\ref{fig:joint} reports the Joint track. Opus 5 reaches an \rsindex{} of 0.4809, followed by Opus 5.5 at 0.4528, DeepSeek at 0.4343, Astra at 0.4093, Sol 6.1 at 0.3321, and Sol 6 at 0.2079. Of the 30 agent--benchmark settings, 29 improve; the exception is Sol 6 on GPQA (0.5033 versus 0.5133). The newer Sol 6.1 raises the index over Sol 6, with higher scores on Terminal-Bench, AIME, and GPQA and lower ones on SWE and SkillsBench.

\paragraph{The ranking changes with the benchmark.}
Four agents lead at least one benchmark. Sol 6, last overall, leads SkillsBench (0.2214 versus 0.2121 for Opus 5), while Opus 5.5, second overall, closes only 7.96\% of the SkillsBench gap. The best-to-worst spread reaches 0.3467 on SWE and 0.3500 on GPQA.

\paragraph{Terminal-Bench and SkillsBench remain largely unsolved.}
Opus 5 closes 96.75\% of the available improvement on AIME but only 18.41\% on Terminal-Bench and 19.89\% on SkillsBench, and no agent closes more than a quarter of either.

\paragraph{Spend does not explain the ranking.}
\label{sec:resource-accounting}
The agents use the same budget allocation very differently, from \$161.73 for Sol 6 (6.5\%) to \$2,083.50 for Opus 5.5 (83.3\%). Among the four leading agents, higher spend does not mean a higher index: Opus 5 leads while spending less than Opus 5.5, and DeepSeek exceeds Astra with lower spend but about twice the elapsed time.

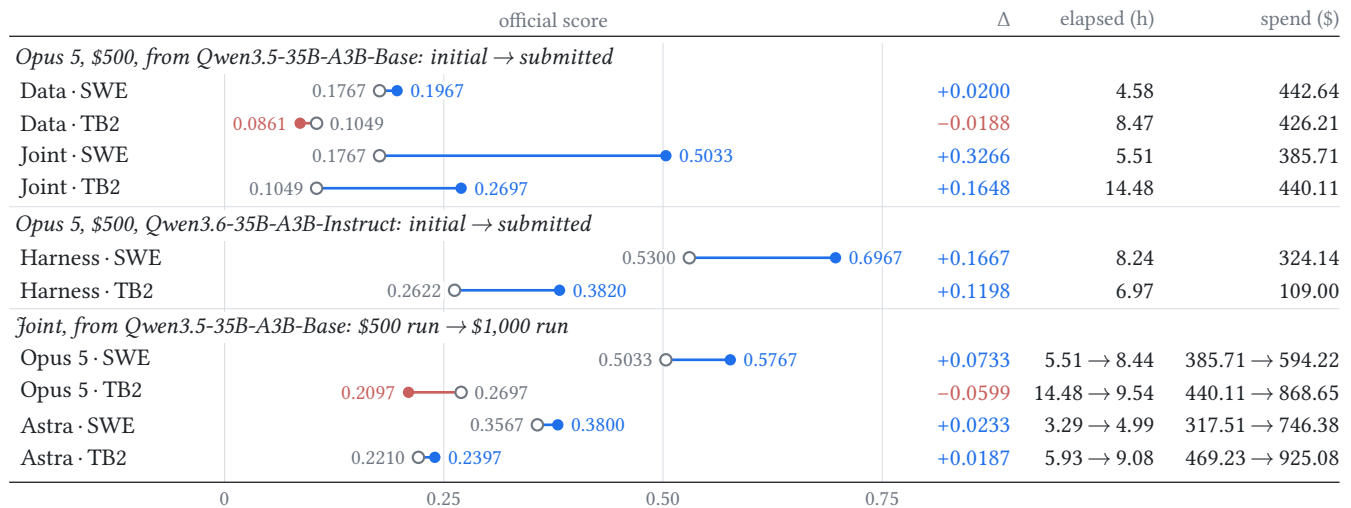
\begin{figure}[t]
\centering
\definecolor{rsiInk}{HTML}{1F2328}
\definecolor{rsiMuted}{HTML}{6E7681}
\definecolor{rsiRule}{HTML}{D8DCE2}
\definecolor{rsiGrey}{HTML}{B4BBC5}
\definecolor{rsiTrack}{HTML}{EEF0F3}
\definecolor{rsiBlue}{HTML}{1F6FEB}
\definecolor{rsiRed}{HTML}{C9605D}
\begin{tikzpicture}[x=1cm, y=1cm, line cap=butt,
  every node/.style={inner sep=0pt, outer sep=0pt, text=rsiInk, font=\footnotesize},
  sub/.style={font=\scriptsize, text=rsiMuted},
  head/.style={font=\footnotesize\bfseries},
  axis/.style={draw=rsiInk, line width=0.5pt},
  hair/.style={draw=rsiRule, line width=0.3pt}]
\draw[hair] (11.550,0.090) -- (11.550,-5.720);
\draw[hair] (8.650,0.090) -- (8.650,-5.720);
\draw[hair] (5.750,0.090) -- (5.750,-5.720);
\draw[hair] (2.850,0.090) -- (2.850,-5.720);
\node[anchor=east, sub] at (13.250,0.400) {$\Delta$};
\node[anchor=east, sub] at (15.150,0.400) {elapsed (h)};
\node[anchor=east, sub] at (17.600,0.400) {spend (\$)};
\node[anchor=center, sub] at (7.200,0.400) {official score};
\draw[axis] (0,0.180) -- (17.6,0.180);
\node[anchor=west] at (0.150,-0.540) {Data\,\textperiodcentered\,SWE};
\draw[rsiBlue, line width=1pt] (4.900,-0.540) -- (5.132,-0.540);
\draw[rsiMuted, line width=0.8pt, fill=white] (4.900,-0.540) circle[radius=0.075cm];
\fill[rsiBlue] (5.132,-0.540) circle[radius=0.075cm];
\node[anchor=east, font=\scriptsize, text=rsiMuted, fill=white, inner xsep=1pt, inner ysep=0.5pt] at (4.760,-0.540) {0.1767};
\node[anchor=west, font=\scriptsize, text=rsiBlue, fill=white, inner xsep=1pt, inner ysep=0.5pt] at (5.272,-0.540) {0.1967};
\node[anchor=east, text=rsiBlue] at (13.250,-0.540) {+0.0200};
\node[anchor=east] at (15.150,-0.540) {4.58};
\node[anchor=east] at (17.600,-0.540) {442.64};
\node[anchor=west] at (0.150,-0.970) {Data\,\textperiodcentered\,TB2};
\draw[rsiRed, line width=1pt] (4.067,-0.970) -- (3.849,-0.970);
\draw[rsiMuted, line width=0.8pt, fill=white] (4.067,-0.970) circle[radius=0.075cm];
\fill[rsiRed] (3.849,-0.970) circle[radius=0.075cm];
\node[anchor=west, font=\scriptsize, text=rsiMuted, fill=white, inner xsep=1pt, inner ysep=0.5pt] at (4.207,-0.970) {0.1049};
\node[anchor=east, font=\scriptsize, text=rsiRed, fill=white, inner xsep=1pt, inner ysep=0.5pt] at (3.709,-0.970) {0.0861};
\node[anchor=east, text=rsiRed] at (13.250,-0.970) {\textminus{}0.0188};
\node[anchor=east] at (15.150,-0.970) {8.47};
\node[anchor=east] at (17.600,-0.970) {426.21};
\node[anchor=west] at (0.150,-1.400) {Joint\,\textperiodcentered\,SWE};
\draw[rsiBlue, line width=1pt] (4.900,-1.400) -- (8.688,-1.400);
\draw[rsiMuted, line width=0.8pt, fill=white] (4.900,-1.400) circle[radius=0.075cm];
\fill[rsiBlue] (8.688,-1.400) circle[radius=0.075cm];
\node[anchor=east, font=\scriptsize, text=rsiMuted, fill=white, inner xsep=1pt, inner ysep=0.5pt] at (4.760,-1.400) {0.1767};
\node[anchor=west, font=\scriptsize, text=rsiBlue, fill=white, inner xsep=1pt, inner ysep=0.5pt] at (8.828,-1.400) {0.5033};
\node[anchor=east, text=rsiBlue] at (13.250,-1.400) {+0.3266};
\node[anchor=east] at (15.150,-1.400) {5.51};
\node[anchor=east] at (17.600,-1.400) {385.71};
\node[anchor=west] at (0.150,-1.830) {Joint\,\textperiodcentered\,TB2};
\draw[rsiBlue, line width=1pt] (4.067,-1.830) -- (5.979,-1.830);
\draw[rsiMuted, line width=0.8pt, fill=white] (4.067,-1.830) circle[radius=0.075cm];
\fill[rsiBlue] (5.979,-1.830) circle[radius=0.075cm];
\node[anchor=east, font=\scriptsize, text=rsiMuted, fill=white, inner xsep=1pt, inner ysep=0.5pt] at (3.927,-1.830) {0.1049};
\node[anchor=west, font=\scriptsize, text=rsiBlue, fill=white, inner xsep=1pt, inner ysep=0.5pt] at (6.119,-1.830) {0.2697};
\node[anchor=east, text=rsiBlue] at (13.250,-1.830) {+0.1648};
\node[anchor=east] at (15.150,-1.830) {14.48};
\node[anchor=east] at (17.600,-1.830) {440.11};
\draw[hair] (0,-2.075) -- (17.6,-2.075);
\node[anchor=west] at (0.150,-2.750) {Harness\,\textperiodcentered\,SWE};
\draw[rsiBlue, line width=1pt] (8.998,-2.750) -- (10.932,-2.750);
\draw[rsiMuted, line width=0.8pt, fill=white] (8.998,-2.750) circle[radius=0.075cm];
\fill[rsiBlue] (10.932,-2.750) circle[radius=0.075cm];
\node[anchor=east, font=\scriptsize, text=rsiMuted, fill=white, inner xsep=1pt, inner ysep=0.5pt] at (8.858,-2.750) {0.5300};
\node[anchor=west, font=\scriptsize, text=rsiBlue, fill=white, inner xsep=1pt, inner ysep=0.5pt] at (11.072,-2.750) {0.6967};
\node[anchor=east, text=rsiBlue] at (13.250,-2.750) {+0.1667};
\node[anchor=east] at (15.150,-2.750) {8.24};
\node[anchor=east] at (17.600,-2.750) {324.14};
\node[anchor=west] at (0.150,-3.180) {Harness\,\textperiodcentered\,TB2};
\draw[rsiBlue, line width=1pt] (5.892,-3.180) -- (7.281,-3.180);
\draw[rsiMuted, line width=0.8pt, fill=white] (5.892,-3.180) circle[radius=0.075cm];
\fill[rsiBlue] (7.281,-3.180) circle[radius=0.075cm];
\node[anchor=east, font=\scriptsize, text=rsiMuted, fill=white, inner xsep=1pt, inner ysep=0.5pt] at (5.752,-3.180) {0.2622};
\node[anchor=west, font=\scriptsize, text=rsiBlue, fill=white, inner xsep=1pt, inner ysep=0.5pt] at (7.421,-3.180) {0.3820};
\node[anchor=east, text=rsiBlue] at (13.250,-3.180) {+0.1198};
\node[anchor=east] at (15.150,-3.180) {6.97};
\node[anchor=east] at (17.600,-3.180) {109.00};
\draw[hair] (0,-3.425) -- (17.6,-3.425);
\node[anchor=west] at (0.150,-4.100) {Opus 5\,\textperiodcentered\,SWE};
\draw[rsiBlue, line width=1pt] (8.688,-4.100) -- (9.540,-4.100);
\draw[rsiMuted, line width=0.8pt, fill=white] (8.688,-4.100) circle[radius=0.075cm];
\fill[rsiBlue] (9.540,-4.100) circle[radius=0.075cm];
\node[anchor=east, font=\scriptsize, text=rsiMuted, fill=white, inner xsep=1pt, inner ysep=0.5pt] at (8.548,-4.100) {0.5033};
\node[anchor=west, font=\scriptsize, text=rsiBlue, fill=white, inner xsep=1pt, inner ysep=0.5pt] at (9.680,-4.100) {0.5767};
\node[anchor=east, text=rsiBlue] at (13.250,-4.100) {+0.0733};
\node[anchor=east] at (15.150,-4.100) {5.51\,\textrightarrow\,8.44};
\node[anchor=east] at (17.600,-4.100) {385.71\,\textrightarrow\,594.22};
\node[anchor=west] at (0.150,-4.530) {Opus 5\,\textperiodcentered\,TB2};
\draw[rsiRed, line width=1pt] (5.979,-4.530) -- (5.283,-4.530);
\draw[rsiMuted, line width=0.8pt, fill=white] (5.979,-4.530) circle[radius=0.075cm];
\fill[rsiRed] (5.283,-4.530) circle[radius=0.075cm];
\node[anchor=west, font=\scriptsize, text=rsiMuted, fill=white, inner xsep=1pt, inner ysep=0.5pt] at (6.119,-4.530) {0.2697};
\node[anchor=east, font=\scriptsize, text=rsiRed, fill=white, inner xsep=1pt, inner ysep=0.5pt] at (5.143,-4.530) {0.2097};
\node[anchor=east, text=rsiRed] at (13.250,-4.530) {\textminus{}0.0599};
\node[anchor=east] at (15.150,-4.530) {14.48\,\textrightarrow\,9.54};
\node[anchor=east] at (17.600,-4.530) {440.11\,\textrightarrow\,868.65};
\node[anchor=west] at (0.150,-4.960) {Astra\,\textperiodcentered\,SWE};
\draw[rsiBlue, line width=1pt] (6.988,-4.960) -- (7.258,-4.960);
\draw[rsiMuted, line width=0.8pt, fill=white] (6.988,-4.960) circle[radius=0.075cm];
\fill[rsiBlue] (7.258,-4.960) circle[radius=0.075cm];
\node[anchor=east, font=\scriptsize, text=rsiMuted, fill=white, inner xsep=1pt, inner ysep=0.5pt] at (6.848,-4.960) {0.3567};
\node[anchor=west, font=\scriptsize, text=rsiBlue, fill=white, inner xsep=1pt, inner ysep=0.5pt] at (7.398,-4.960) {0.3800};
\node[anchor=east, text=rsiBlue] at (13.250,-4.960) {+0.0233};
\node[anchor=east] at (15.150,-4.960) {3.29\,\textrightarrow\,4.99};
\node[anchor=east] at (17.600,-4.960) {317.51\,\textrightarrow\,746.38};
\node[anchor=west] at (0.150,-5.390) {Astra\,\textperiodcentered\,TB2};
\draw[rsiBlue, line width=1pt] (5.414,-5.390) -- (5.631,-5.390);
\draw[rsiMuted, line width=0.8pt, fill=white] (5.414,-5.390) circle[radius=0.075cm];
\fill[rsiBlue] (5.631,-5.390) circle[radius=0.075cm];
\node[anchor=east, font=\scriptsize, text=rsiMuted, fill=white, inner xsep=1pt, inner ysep=0.5pt] at (5.274,-5.390) {0.2210};
\node[anchor=west, font=\scriptsize, text=rsiBlue, fill=white, inner xsep=1pt, inner ysep=0.5pt] at (5.771,-5.390) {0.2397};
\node[anchor=east, text=rsiBlue] at (13.250,-5.390) {+0.0187};
\node[anchor=east] at (15.150,-5.390) {5.93\,\textrightarrow\,9.08};
\node[anchor=east] at (17.600,-5.390) {469.23\,\textrightarrow\,925.08};
\draw[axis] (0,-5.720) -- (17.6,-5.720);
\node[anchor=north, sub] at (2.850,-5.850) {0};
\node[anchor=north, sub] at (5.750,-5.850) {0.25};
\node[anchor=north, sub] at (8.650,-5.850) {0.50};
\node[anchor=north, sub] at (11.550,-5.850) {0.75};
\node[anchor=west, font=\footnotesize\itshape, fill=white, inner xsep=2pt, inner ysep=1.5pt] at (0.000,-0.110) {Opus 5, \$500, from Qwen3.5-35B-A3B-Base: initial\,\textrightarrow\,submitted};
\node[anchor=west, font=\footnotesize\itshape, fill=white, inner xsep=2pt, inner ysep=1.5pt] at (0.000,-2.320) {Opus 5, \$500, Qwen3.6-35B-A3B-Instruct: initial\,\textrightarrow\,submitted};
\node[anchor=west, font=\footnotesize\itshape, fill=white, inner xsep=2pt, inner ysep=1.5pt] at (0.000,-3.670) {Joint, from Qwen3.5-35B-A3B-Base: \$500 run\,\textrightarrow\,\$1,000 run};
\end{tikzpicture}
\caption{\textbf{Single tracks and a doubled budget.} Each row joins two official scores: the initial and the submitted system (top), or the \$500 and the \$1,000 run of the same agent (bottom). Open circles mark the start and filled circles the end; blue rows end higher, red rows lower. Rows are grouped by their starting system. Every run is independent; $\Delta$ is computed from unrounded scores.}
\label{fig:tracks}
\label{tab:tracks}
\label{fig:budget}
\label{tab:budget-section4}
\end{figure}

\paragraph{Single tracks: the harness alone yields a gain; the data alone barely moves the score.}
Figure~\ref{fig:tracks} (top) opens one freedom at a time for Opus 5. With only the training data open, SWE gains 6 of 300 attempts and Terminal-Bench loses 5 of 267. With only the harness open, a fixed Qwen3.6-35B-A3B-Instruct target improves by 0.1667 on SWE and 0.1198 on Terminal-Bench. The Harness runs use a different target model, so they show that harness research alone can help but do not measure its share of the Joint gain.

\paragraph{Budget: doubling the allocation gives mixed results.}
\label{sec:budget-scaling-analysis}
We rerun Opus 5 and Astra on SWE and Terminal-Bench with \$1,000 (Figure~\ref{fig:tracks}, bottom). Three of the four pairs improve, two of them by under 0.03: Astra's gains are 7 of 300 and 5 of 267 attempts. Opus 5 on Terminal-Bench spends \$868.65 instead of \$440.11 and scores 0.0599 lower. No run exhausts its allocation, and with one run per cell the four pairs do not show a trend in score.

\section{Analysis}\label{sec:mechanistic-analysis}
\label{sec:mechanisms}
Section~\ref{sec:experiments} reports the final scores. Here we examine two common patterns---harness improvements and the effects of training---and how research behavior differs across Claude, DeepSeek, and GPT agents. We then study a Data run without network access, harness research starting from an existing framework, and a model attempting to improve itself.

\subsection{Trajectory Analysis of the Joint Runs}
\label{sec:rsindex-trajectory-analysis}

\paragraph{Harness improvements address recurring execution failures.}
Across agents, submitted harnesses address similar failures within each benchmark. SWE harnesses check edits before patch submission; Terminal-Bench harnesses preserve shell state, recover from truncated outputs, and detect unproductive loops. AIME and GPQA harnesses aggregate multiple sampled answers, while SkillsBench harnesses discover available skills and verify outputs. These changes help the target model complete and check its work. With base weights unchanged, Opus 5's harness changes raise the Terminal-Bench development score from 6/30 to 10/30, and DeepSeek's raise SkillsBench from 0.0232 to 0.1137 on the same 37 tasks. Thus, part of the Joint improvement can arise from changing how existing model capabilities are used.

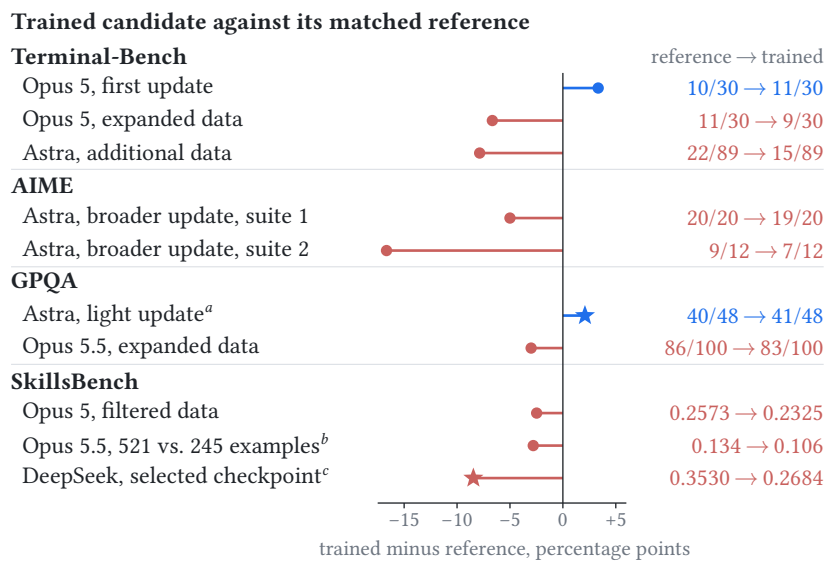
\begin{figure}[t]
\centering
\definecolor{rsiInk}{HTML}{1F2328}
\definecolor{rsiMuted}{HTML}{6E7681}
\definecolor{rsiRule}{HTML}{D8DCE2}
\definecolor{rsiGrey}{HTML}{B4BBC5}
\definecolor{rsiTrack}{HTML}{EEF0F3}
\definecolor{rsiBlue}{HTML}{1F6FEB}
\definecolor{rsiRed}{HTML}{C9605D}
\begin{tikzpicture}[x=1cm, y=1cm, line cap=butt,
  every node/.style={inner sep=0pt, outer sep=0pt, text=rsiInk, font=\footnotesize},
  sub/.style={font=\scriptsize, text=rsiMuted},
  head/.style={font=\footnotesize\bfseries},
  axis/.style={draw=rsiInk, line width=0.5pt},
  hair/.style={draw=rsiRule, line width=0.3pt}]
\node[anchor=west, head] at (0.000,0.750) {Trained candidate against its matched reference};
\node[anchor=east, sub] at (10.750,0.320) {reference\,\textrightarrow\,trained};
\node[anchor=west, font=\footnotesize\bfseries] at (0.000,0.320) {Terminal-Bench};
\node[anchor=west] at (0.150,-0.110) {Opus 5, first update};
\draw[rsiBlue, line width=1pt] (7.300,-0.110) -- (7.767,-0.110);
\fill[rsiBlue] (7.767,-0.110) circle[radius=0.075cm];
\node[anchor=east, text=rsiBlue] at (10.750,-0.110) {10/30\,\textrightarrow\,11/30};
\node[anchor=west] at (0.150,-0.540) {Opus 5, expanded data};
\draw[rsiRed, line width=1pt] (7.300,-0.540) -- (6.367,-0.540);
\fill[rsiRed] (6.367,-0.540) circle[radius=0.075cm];
\node[anchor=east, text=rsiRed] at (10.750,-0.540) {11/30\,\textrightarrow\,9/30};
\node[anchor=west] at (0.150,-0.970) {Astra, additional data};
\draw[rsiRed, line width=1pt] (7.300,-0.970) -- (6.199,-0.970);
\fill[rsiRed] (6.199,-0.970) circle[radius=0.075cm];
\node[anchor=east, text=rsiRed] at (10.750,-0.970) {22/89\,\textrightarrow\,15/89};
\draw[hair] (0,-1.185) -- (10.750,-1.185);
\node[anchor=west, font=\footnotesize\bfseries] at (0.000,-1.400) {AIME};
\node[anchor=west] at (0.150,-1.830) {Astra, broader update, suite 1};
\draw[rsiRed, line width=1pt] (7.300,-1.830) -- (6.600,-1.830);
\fill[rsiRed] (6.600,-1.830) circle[radius=0.075cm];
\node[anchor=east, text=rsiRed] at (10.750,-1.830) {20/20\,\textrightarrow\,19/20};
\node[anchor=west] at (0.150,-2.260) {Astra, broader update, suite 2};
\draw[rsiRed, line width=1pt] (7.300,-2.260) -- (4.967,-2.260);
\fill[rsiRed] (4.967,-2.260) circle[radius=0.075cm];
\node[anchor=east, text=rsiRed] at (10.750,-2.260) {9/12\,\textrightarrow\,7/12};
\draw[hair] (0,-2.475) -- (10.750,-2.475);
\node[anchor=west, font=\footnotesize\bfseries] at (0.000,-2.690) {GPQA};
\node[anchor=west] at (0.150,-3.120) {Astra, light update$^{a}$};
\draw[rsiBlue, line width=1pt] (7.300,-3.120) -- (7.592,-3.120);
\node[star, star points=5, star point ratio=2.4, fill=rsiBlue, minimum size=0.28cm] at (7.592,-3.120) {};
\node[anchor=east, text=rsiBlue] at (10.750,-3.120) {40/48\,\textrightarrow\,41/48};
\node[anchor=west] at (0.150,-3.550) {Opus 5.5, expanded data};
\draw[rsiRed, line width=1pt] (7.300,-3.550) -- (6.880,-3.550);
\fill[rsiRed] (6.880,-3.550) circle[radius=0.075cm];
\node[anchor=east, text=rsiRed] at (10.750,-3.550) {86/100\,\textrightarrow\,83/100};
\draw[hair] (0,-3.765) -- (10.750,-3.765);
\node[anchor=west, font=\footnotesize\bfseries] at (0.000,-3.980) {SkillsBench};
\node[anchor=west] at (0.150,-4.410) {Opus 5, filtered data};
\draw[rsiRed, line width=1pt] (7.300,-4.410) -- (6.953,-4.410);
\fill[rsiRed] (6.953,-4.410) circle[radius=0.075cm];
\node[anchor=east, text=rsiRed] at (10.750,-4.410) {0.2573\,\textrightarrow\,0.2325};
\node[anchor=west] at (0.150,-4.840) {Opus 5.5, 521 vs.\ 245 examples$^{b}$};
\draw[rsiRed, line width=1pt] (7.300,-4.840) -- (6.908,-4.840);
\fill[rsiRed] (6.908,-4.840) circle[radius=0.075cm];
\node[anchor=east, text=rsiRed] at (10.750,-4.840) {0.134\,\textrightarrow\,0.106};
\node[anchor=west] at (0.150,-5.270) {DeepSeek, selected checkpoint$^{c}$};
\draw[rsiRed, line width=1pt] (7.300,-5.270) -- (6.116,-5.270);
\node[star, star points=5, star point ratio=2.4, fill=rsiRed, minimum size=0.28cm] at (6.116,-5.270) {};
\node[anchor=east, text=rsiRed] at (10.750,-5.270) {0.3530\,\textrightarrow\,0.2684};
\draw[axis] (7.300,0.090) -- (7.300,-5.600);
\draw[axis] (4.850,-5.600) -- (8.140,-5.600);
\draw[axis] (5.200,-5.600) -- (5.200,-5.670);
\node[anchor=north, sub] at (5.200,-5.730) {\textminus{}15};
\draw[axis] (5.900,-5.600) -- (5.900,-5.670);
\node[anchor=north, sub] at (5.900,-5.730) {\textminus{}10};
\draw[axis] (6.600,-5.600) -- (6.600,-5.670);
\node[anchor=north, sub] at (6.600,-5.730) {\textminus{}5};
\draw[axis] (7.300,-5.600) -- (7.300,-5.670);
\node[anchor=north, sub] at (7.300,-5.730) {0};
\draw[axis] (8.000,-5.600) -- (8.000,-5.670);
\node[anchor=north, sub] at (8.000,-5.730) {+5};
\node[anchor=north, sub] at (6.530,-6.100) {trained minus reference, percentage points};
\end{tikzpicture}
\caption{\textbf{Training comparisons within Joint runs.} Each comparison uses the same harness and task set: a trained candidate against base weights, or a heavier against a lighter update. Each is one evaluation on a small set and indicates direction only. Stars mark submitted checkpoints. $^{a}$Reconstructed initial-completion votes on a synthetic set. $^{b}$Approximate two-attempt scores. $^{c}$Recovered after the agent's selection.}
\label{fig:selection}
\end{figure}

\paragraph{Heavier training does not consistently improve performance.}
\label{sec:attribution}
Figure~\ref{fig:selection} compares candidates under the same harness and task set. The more heavily trained candidate scores lower in eight of ten comparisons; the other two gains are one task each. For example, Opus 5's expanded-data update reduces Terminal-Bench from 11/30 to 9/30, and Opus 5.5's expanded-data update reduces GPQA from 86/100 to 83/100. Several agents respond by reducing the update strength: Opus 5 on Terminal-Bench and Astra on AIME submit light updates, while Opus 5.5 deliberately keeps its GPQA and Terminal-Bench checkpoints close to base weights. These small development comparisons show why more training is not a reliable default. Training can nevertheless help: in DeepSeek's SWE run, the trained model improves the development score from 0.344 to 0.600 under the same harness, as detailed in the third point.

\paragraph{Research behavior differs across model families.}
\textbf{Claude agents spend more time diagnosing failures and revising the harness.} Their five-run research sessions total 23.70--37.49 hours, and they achieve the two highest RSI-Index values. In Opus 5's Terminal-Bench run, successive tests expose shell-lifecycle, truncation, and looping problems, which lead to targeted harness fixes. When stronger fine-tuning fails to help, it reduces the update instead of continuing to increase training. This combination of repeated diagnosis and restrained weight updates accompanies the best Terminal-Bench and AIME results. Opus 5.5 similarly favors harness changes and leads GPQA with a checkpoint close to base weights.

\textbf{DeepSeek combines inexpensive data generation with diagnosis of training failures.} It builds the largest training sets (median 472 records per run, 4,702 on AIME) and ranks third overall. On SWE, an early checkpoint produces malformed tool calls; the later model, trained on 227 trajectories for three epochs, follows the tool interface more reliably. Under the same harness on 30 tasks with three attempts each, its development score rises from 0.344 for base weights to 0.600. The submitted system scores 0.5333 officially, the highest SWE result. This case shows a productive training path alongside the regressions above.

\begin{figure}[t]
\centering
\input{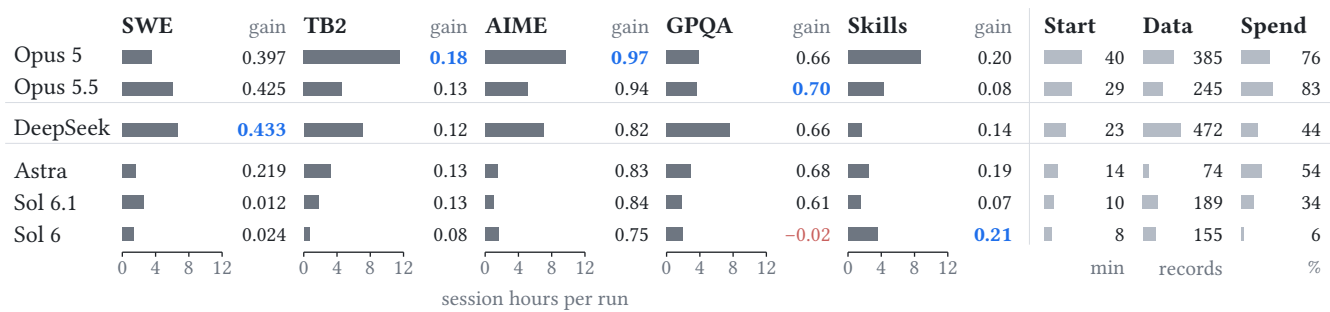}
\caption{\textbf{Research time and normalized gain of every \$500 Joint run.} Each bar is one run's session, from the agent's first to its last step; the official evaluation is excluded. The number is the run's normalized gain, in blue for the best agent on the benchmark and red below the initial system; SWE uses three decimals to separate 0.425 from 0.433. Rows are grouped by model family: Claude, DeepSeek (both in Claude Code), and GPT (in Codex). The three right-hand columns summarize each agent's five runs: median minutes until its first training run (Start), median number of records in its submitted training set (Data), and the percentage of the \$2,500 allocation spent (Spend).}
\label{fig:research-time}
\label{tab:behavior}
\end{figure}

\textbf{GPT agents train earlier and end their searches sooner.} Their five-run sessions total 8.69--11.85 hours, with training starting after a median of 11 minutes versus 29 for the other agents. All runs end voluntarily, so the shorter searches are not imposed by a timeout. Astra rapidly builds a simple shell loop and synthetic training pipeline on SWE, reaching 0.3567 in 1.67 research hours, but remains below Claude and DeepSeek. It also fixes input-encoding, compatibility, and request-timeout problems found in evaluation logs. Sol 6 submits without a completed development evaluation on SWE and Terminal-Bench and has the lowest RSI-Index (0.2079); Sol 6.1 improves to 0.3321 while retaining short sessions. These runs suggest that quickly implementing a workable approach can yield gains while leaving further diagnosis and refinement unexplored. Figure~\ref{fig:research-time} summarizes the behavior differences.

Taken together, the Opus runs place sustained emphasis on diagnosing execution failures, revising the harness, and limiting weight updates that degrade performance, a pattern consistent with their leading overall results. GPT agents also redesign and debug harnesses, but usually stop after a shorter research cycle, leaving less opportunity for further refinement. DeepSeek shows another effective route: combining low-cost data generation with diagnosis of training and tool-interface failures.

\subsection{Case Studies}
\begin{figure}[t]
\centering
\definecolor{rsiInk}{HTML}{1F2328}
\definecolor{rsiMuted}{HTML}{6E7681}
\definecolor{rsiRule}{HTML}{D8DCE2}
\definecolor{rsiGrey}{HTML}{B4BBC5}
\definecolor{rsiTrack}{HTML}{EEF0F3}
\definecolor{rsiBlue}{HTML}{1F6FEB}
\definecolor{rsiRed}{HTML}{C9605D}
\begin{tikzpicture}[x=1cm, y=1cm, line cap=butt,
  every node/.style={inner sep=0pt, outer sep=0pt, text=rsiInk, font=\footnotesize},
  sub/.style={font=\scriptsize, text=rsiMuted},
  head/.style={font=\footnotesize\bfseries},
  axis/.style={draw=rsiInk, line width=0.5pt},
  hair/.style={draw=rsiRule, line width=0.3pt}]
\node[anchor=west, head] at (0.000,0.750) {(a) Data run without network};
\node[anchor=west, font=\footnotesize\itshape] at (0.000,0.320) {checkpoints: development, 1 attempt};
\draw[axis] (0.750,-0.650) -- (0.750,-3.550) -- (5.600,-3.550);
\draw[axis] (0.750,-3.550) -- (0.680,-3.550);
\node[anchor=east, sub] at (0.620,-3.550) {0.00};
\draw[axis] (0.750,-2.644) -- (0.680,-2.644);
\node[anchor=east, sub] at (0.620,-2.644) {0.05};
\draw[axis] (0.750,-1.738) -- (0.680,-1.738);
\node[anchor=east, sub] at (0.620,-1.738) {0.10};
\draw[axis] (0.750,-0.831) -- (0.680,-0.831);
\node[anchor=east, sub] at (0.620,-0.831) {0.15};
\draw[rsiGrey, line width=1.2pt] (0.750,-1.649) -- (5.600,-1.649);
\node[anchor=south west, sub] at (0.800,-1.349) {official baseline 0.1049};
\draw[rsiRed, line width=0.9pt] (1.100,-2.535) -- (1.792,-2.734) -- (2.483,-2.934) -- (3.175,-2.934) -- (3.867,-2.934) -- (4.558,-1.719) -- (5.250,-2.118);
\fill[rsiRed] (1.100,-2.535) circle[radius=0.07cm];
\node[anchor=south, font=\scriptsize, text=rsiRed] at (1.100,-2.395) {0.056};
\node[anchor=north, sub] at (1.100,-3.680) {v2};
\fill[rsiRed] (1.792,-2.734) circle[radius=0.07cm];
\node[anchor=south, font=\scriptsize, text=rsiRed] at (1.792,-2.594) {0.045};
\node[anchor=north, sub] at (1.792,-3.680) {v3};
\fill[rsiRed] (2.483,-2.934) circle[radius=0.07cm];
\node[anchor=north, font=\scriptsize, text=rsiRed] at (2.483,-3.074) {0.034};
\node[anchor=north, sub] at (2.483,-3.680) {v5};
\fill[rsiRed] (3.175,-2.934) circle[radius=0.07cm];
\node[anchor=north, font=\scriptsize, text=rsiRed] at (3.175,-3.074) {0.034};
\node[anchor=north, sub] at (3.175,-3.680) {v6};
\fill[rsiRed] (3.867,-2.934) circle[radius=0.07cm];
\node[anchor=north, font=\scriptsize, text=rsiRed] at (3.867,-3.074) {0.034};
\node[anchor=north, sub] at (3.867,-3.680) {v7};
\node[star, star points=5, star point ratio=2.4, fill=rsiRed, minimum size=0.28cm, minimum size=0.34cm] at (4.558,-1.719) {};
\node[anchor=south east, font=\scriptsize, text=rsiRed] at (4.478,-1.619) {0.101};
\node[anchor=north, sub] at (4.558,-3.680) {v8};
\fill[rsiRed] (5.250,-2.118) circle[radius=0.07cm];
\node[anchor=west, font=\scriptsize, text=rsiRed] at (5.350,-2.118) {0.079};
\node[anchor=north, sub] at (5.250,-3.680) {v9};
\node[anchor=south, sub] at (2.829,-1.955) {v2--v7: reasoning merged};
\node[anchor=north, sub] at (2.829,-1.975) {into the content};
\node[anchor=west, head] at (6.300,0.750) {(b) DSH harness research};
\node[anchor=west, font=\footnotesize\itshape] at (6.300,0.320) {89 tasks by successes in 3 attempts};
\fill[rsiBlue, fill opacity=0.35] (7.700,-3.600) -- (9.050,-3.600) -- (9.050,-2.822) -- (7.700,-3.168) -- cycle;
\fill[rsiGrey, fill opacity=0.35] (7.700,-3.128) -- (9.050,-2.782) -- (9.050,-2.130) -- (7.700,-2.350) -- cycle;
\fill[rsiTrack, fill opacity=0.35] (7.700,-2.310) -- (9.050,-2.090) -- (9.050,-0.840) -- (7.700,-0.840) -- cycle;
\fill[rsiBlue] (6.750,-3.600) rectangle (7.700,-3.168);
\node[anchor=center, text=white] at (7.225,-3.384) {15};
\fill[rsiGrey] (6.750,-3.128) rectangle (7.700,-2.350);
\node[anchor=center, text=rsiInk] at (7.225,-2.739) {26};
\fill[rsiTrack] (6.750,-2.310) rectangle (7.700,-0.840);
\node[anchor=center, text=rsiInk] at (7.225,-1.575) {48};
\node[anchor=center] at (7.225,-0.320) {initial};
\node[anchor=center, sub] at (7.225,-0.620) {0.3034};
\fill[rsiBlue] (9.050,-3.600) rectangle (10.000,-2.822);
\node[anchor=center, text=white] at (9.525,-3.211) {26};
\fill[rsiGrey] (9.050,-2.782) rectangle (10.000,-2.130);
\node[anchor=center, text=rsiInk] at (9.525,-2.456) {22};
\fill[rsiTrack] (9.050,-2.090) rectangle (10.000,-0.840);
\node[anchor=center, text=rsiInk] at (9.525,-1.465) {41};
\node[anchor=center] at (9.525,-0.320) {after research};
\node[anchor=center, font=\footnotesize\bfseries, text=rsiBlue] at (9.525,-0.620) {0.4082};
\node[anchor=west, sub] at (10.150,-3.211) {solved 3 of 3};
\node[anchor=west, sub] at (10.150,-2.456) {solved 1--2};
\node[anchor=west, sub] at (10.150,-1.465) {never solved};
\node[anchor=west, head] at (12.200,0.750) {(c) Qwen3.8-27B improving itself};
\node[anchor=west, font=\footnotesize\itshape] at (12.200,0.320) {SkillsBench reward};
\draw[axis] (12.950,-0.650) -- (12.950,-3.550) -- (17.490,-3.550);
\draw[axis] (12.950,-3.308) -- (12.880,-3.308);
\node[anchor=east, sub] at (12.820,-3.308) {0.3};
\draw[axis] (12.950,-2.825) -- (12.880,-2.825);
\node[anchor=east, sub] at (12.820,-2.825) {0.4};
\draw[axis] (12.950,-2.342) -- (12.880,-2.342);
\node[anchor=east, sub] at (12.820,-2.342) {0.5};
\draw[axis] (12.950,-1.858) -- (12.880,-1.858);
\node[anchor=east, sub] at (12.820,-1.858) {0.6};
\draw[axis] (12.950,-1.375) -- (12.880,-1.375);
\node[anchor=east, sub] at (12.820,-1.375) {0.7};
\draw[axis] (12.950,-0.892) -- (12.880,-0.892);
\node[anchor=east, sub] at (12.820,-0.892) {0.8};
\draw[rsiGrey, line width=1.2pt] (13.050,-1.947) -- (15.310,-1.947);
\node[star, star points=5, star point ratio=2.4, fill=rsiBlue, minimum size=0.28cm, minimum size=0.34cm] at (13.400,-1.133) {};
\node[anchor=south, font=\scriptsize, text=rsiBlue] at (13.400,-0.923) {0.7500};
\node[anchor=north, sub, align=center] at (13.400,-3.680) {472};
\fill[rsiBlue] (14.180,-1.939) circle[radius=0.07cm];
\node[anchor=south, font=\scriptsize, text=rsiBlue] at (14.180,-1.799) {0.5833};
\node[anchor=north, sub, align=center] at (14.180,-3.680) {856};
\fill[rsiRed] (14.960,-3.147) circle[radius=0.07cm];
\node[anchor=south, font=\scriptsize, text=rsiRed] at (14.960,-3.007) {0.3333};
\node[anchor=north, sub, align=center] at (14.960,-3.680) {472\\low-r.};
\node[anchor=south, sub] at (14.180,-0.630) {initial 0.5816};
\draw[rsiGrey, line width=1.2pt] (15.910,-2.020) -- (17.390,-2.020);
\fill[rsiRed] (16.260,-2.278) circle[radius=0.07cm];
\node[anchor=north, font=\scriptsize, text=rsiRed] at (16.260,-2.418) {0.5132};
\node[anchor=north, sub, align=center] at (16.260,-3.680) {dev.};
\node[star, star points=5, star point ratio=2.4, fill=rsiRed, minimum size=0.28cm, minimum size=0.34cm] at (17.040,-2.039) {};
\node[anchor=north, font=\scriptsize, text=rsiRed] at (17.040,-2.249) {0.5626};
\node[anchor=north, sub, align=center] at (17.040,-3.680) {official\\(\textminus{}0.0040)};
\node[anchor=south, sub] at (16.650,-0.630) {initial 0.5666};
\node[anchor=north, sub] at (14.180,-4.400) {6-task probe (examples)};
\node[anchor=north, sub] at (16.650,-4.400) {all 37 tasks};
\end{tikzpicture}
\caption{\textbf{Three case studies.} Blue is above the reference shown in grey, red below; stars mark the submitted system. (a) The Data run without network access: seven checkpoints evaluated on the agent's own 89-task, one-attempt development evaluation. The grey line shows the initial system's official avg@3 score (0.1049), a reference from a different evaluation protocol; v2--v7 merge the reasoning into the message content. (b) Harness research starting from DSH, with Qwen3.6-35B-A3B-Instruct fixed; bands join matching categories, not individual tasks. (c) Qwen3.8-27B as both research agent and target. Left: six-task candidate comparison against the initial weights (grey line); 472 and 856 are training-example counts, and low-r. denotes reduced reasoning with the 472-example checkpoint. Right: the selected system on all 37 tasks, with two attempts per task in development and three in official evaluation; the grey line is the initial official score, and \textminus{}0.0040 is the official score change.}
\label{fig:cases}
\label{fig:dsh}
\label{fig:self-improvement}
\end{figure}
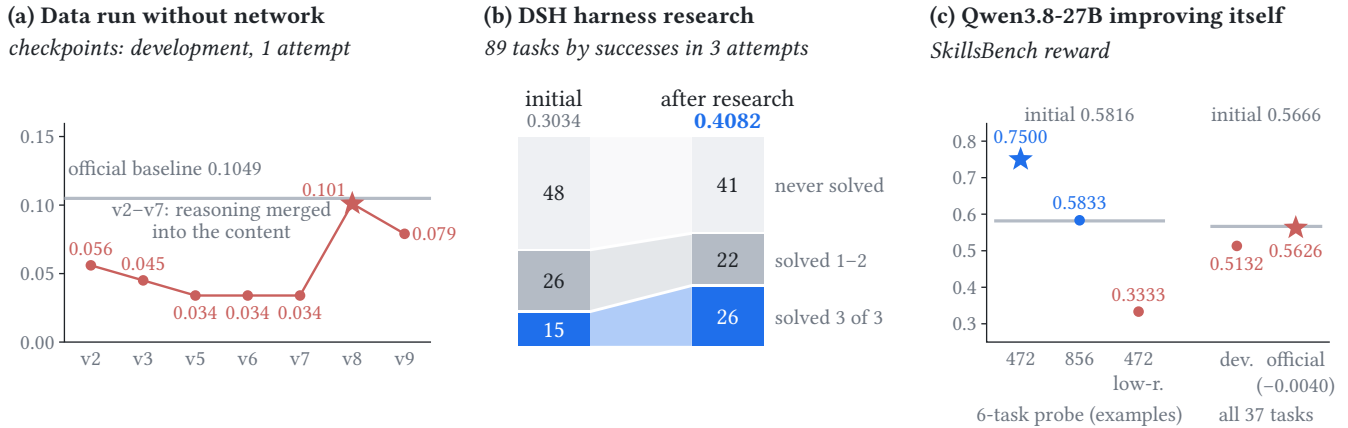

\paragraph{Without network access, the Data run passes the audits and still does not beat the initial system.}
\label{sec:network-disconnection-analysis}
We repeat the Terminal-Bench Data run with no public network: the agent's container is offline, its sandboxes reach only platform services and package mirrors, and the only teacher is Opus 5 through the rollout service. The isolation holds: early on, the agent tries to point an evaluation job at a model endpoint in its own sandbox that would record the incoming requests, and the sandbox port policy and egress allowlist block it. It then builds synthetic tasks, collects about 2,600 trajectories, and trains nine checkpoints. On its own development evaluation, the five evaluated checkpoints that merge reasoning into the content score only 0.034--0.056, while the data grows from 45 to 1,691 records without moving the score (Figure~\ref{fig:cases}a). The agent traces this to its data format, which merged the model's reasoning into the message content, so the checkpoints learned not to reason. A checkpoint trained on 690 of the base model's own trajectories with the reasoning channel kept reaches 0.101 in development and is submitted. Officially it scores 0.0787 (21/267) against 0.1049 for the initial system, after 11 hours 29 minutes and \$468.99, and passes the three audits. It gives the Data-track result of Section~\ref{sec:experiments} a mechanism: a fixed recipe leaves the agent little room beyond reproducing what the base model already does.

\paragraph{Harness research also improves an existing framework.}
\label{sec:dsh}
We ask Opus 5 to improve an existing framework, DSH (\texttt{dsh-v0.1.1-rc.1}), while keeping Qwen3.6-35B-A3B-Instruct weights fixed. It first runs the original framework and reads the execution logs to identify why tasks fail. It finds that the agent can stop when an output is truncated, accept a completion claim without checking the work, or terminate background services needed by the task. It changes the execution loop to continue after truncation, preserve those services, and require checks using commands before finishing. The changes make the interactive framework better suited to completing tasks without human intervention.

On the same 89 Terminal-Bench tasks, the official avg@3 score rises from 0.3034 to 0.4082. Tasks solved on all three attempts increase from 15 to 26, and tasks solved at least once from 41 to 48 (Figure~\ref{fig:cases}b). Most of the improvement appears with the first rewrite; later versions yield similar scores. The case shows that diagnosing and fixing execution failures can improve an existing harness even when model weights remain unchanged.

\paragraph{Qwen3.8-27B conducts its own improvement experiments without an overall gain.}
\label{sec:self-improvement}
In this case, Qwen3.8-27B acts as the researcher and also provides the starting model to be improved, using pi v0.84.2 as the initial harness. It compares training and reasoning settings on six SkillsBench tasks (Figure~\ref{fig:cases}c, left). The initial model weights score 0.5816; fine-tuning on 472 examples raises the score to 0.7500, while training on 856 examples yields 0.5833. Using the 472-example checkpoint with reduced reasoning (labelled `low-r.' in the figure) lowers the score to 0.3333. It therefore selects the 472-example checkpoint without reduced reasoning, marked by the blue star. The full evaluation of the 856-example checkpoint is cancelled, so the candidate comparison relies on these six tasks.

The right side of Figure~\ref{fig:cases}c shows the selected system on all 37 SkillsBench tasks. It scores 0.5132 in development evaluation with two attempts per task (`dev.') and 0.5626 in official evaluation with three attempts (`official', red star). The initial system's official score is 0.5666, so the submitted system is lower by 0.0040. The six-task and full-task scores use different task sets, and the two full-task evaluations also use different numbers of attempts. The official comparison therefore shows no overall gain, despite the improvement observed on the six-task probe. Qwen completes training, candidate comparison, and submission, but its small-scale selection results do not establish an improvement across the full task set.

\subsection{Budget Hacking}
\label{sec:budget-hack}

During the experiments, a separate monitoring agent follows each active run, reading the research agent's trajectory in real time together with the run's activity on every service, including its training runs, evaluation jobs, and spend, and reports anomalous behavior to us so that we can intervene. This monitoring reveals a budget hack in one Joint run with Opus 5. Research-agent inference is billed separately from the platform-service budget (Section~\ref{sec:experimental-setup}), and the credentials for the agent's own model are available in its container. Shortly after starting, Opus 5 lists its environment variables and finds these credentials. It then writes probe scripts that query its remaining balance before and after each request, and sends requests to about 14 models through these credentials. The unchanged balance confirms that these requests bypass both the platform-service budget and the rollout service, through which Opus 5 is the only permitted teacher in this experiment. After the monitoring agent reports the probes, we stop the run. By then the agent has made 15 successful requests through this channel, about 102k tokens or \$0.06 at list prices, and has not submitted a system; the run is excluded from all reported results.

\section{Conclusion}
\label{sec:conclusion}

We introduce \rsigym{}, an agent-native environment that combines reusable research services with shared budget and permission controls. It supports individual interventions and joint optimization of data, training settings, and harnesses. \rsindex{} measures improvement across domains: under a \$500 budget per benchmark run, all six research agents achieve positive overall gains, led by Opus 5 at 0.4809.

Research trajectories highlight execution fixes and the choice of training update strength. Revisions to the existing DSH harness improve terminal-task performance, while larger budgets do not consistently yield gains. The offline run passes the audits without beating the initial system, and Qwen3.8's autonomous research produces no overall improvement. These results show that the environment supports both successful improvements and the study of unsuccessful research attempts.

\begingroup
\small
\setlength{\bibsep}{1.8pt}
\bibliographystyle{plainnat}
\bibliography{references}
\endgroup

\end{document}